\documentclass[11pt]{article}
 \usepackage{benstyle}
 
\usepackage{longtable}

\usepackage{cite}

\newtheorem{assumption}[theorem]{Assumption}

\title{Optimal sampling for least-squares approximation}
\author{Ben Adcock \\ Department of Mathematics \\ Simon Fraser University \\ Canada}

\begin{document}

\maketitle

\begin{abstract}
Least-squares approximation is one of the most important methods for recovering an unknown function from data. While in many applications the data is fixed, in many others there is substantial freedom to choose where to sample. In this paper, we review recent progress on near-optimal random sampling strategies for (weighted) least-squares approximation in arbitrary linear spaces. We introduce the Christoffel function as a key quantity in the analysis of (weighted) least-squares approximation from random samples, then show how it can be used to construct a random sampling strategy, termed \textit{Christoffel sampling}, that possesses near-optimal sample complexity: namely, the number of samples scales log-linearly in the dimension of the approximation space $n$. We discuss a series of variations, extensions and further topics, and throughout highlight connections to approximation theory, machine learning, information-based complexity and numerical linear algebra. Finally, motivated by various contemporary applications, we consider a generalization of the classical setting where the samples need not be pointwise samples of a scalar-valued function, and the approximation space need not be linear. We show that, even in this significantly more general setting, suitable generalizations of Christoffel function still determine the sample complexity. Consequently, these can be used to design enhanced, Christoffel sampling strategies in a unified way for general recovery problems. This article is largely self-contained, and intended to be accessible to nonspecialists.
\end{abstract}

\section{Introduction}

Least-squares approximation is the process of recovering an unknown function from samples by computing a best $\ell^2$-norm fit to the data in a given subspace -- often termed the \textit{approximation space}. This is a classical approach, yet it is one of the widely-used tools in applied mathematics, computer science, engineering and numerous other disciplines. For the data scientist, it is almost always the first `go-to' method when trying to fit a function to data.

In many data-fitting problems, the sample points are fixed. However, many other problems offer substantial flexibility to choose where to sample. When data is also expensive to acquire -- which, despite claims about `big data' is often the case in applications in science and engineering -- we are naturally led to the following questions. How many samples do we need -- or, in other words, what is the \textit{sample complexity} -- and how should we best choose them? This is by no means a new question. It arises in many different guises in different fields, including \textit{optimal design of experiments} in statistics, \textit{active learning} in machine learning, \textit{optimal sensor placement} in sampling theory and signal processing, and \textit{optimal (standard) information} in information-based complexity.

The purpose of this article is to survey recent advances made in the last 5-10 years in \textit{optimal sampling}, as we shall term it from now on, which has been motivated by certain function approximation problems in high dimensions. Throughout, our aim is to establish quasi-optimal recovery (in an appropriate sense) with near-optimal sample complexity.  The approach we describe, which we term \textit{Christoffel sampling}, is, in essence, an importance sampling technique, where samples are drawn randomly from a probability measure chosen specifically for the given approximation space. As we highlight, this approach is elegantly simple, broadly applicable and, in many cases, straightforward to implement numerically.

\subsection{Overview}

After a short literature review (\S \ref{s:lit-rev}), this article commences with a formulation and summary of (weighted) least-squares approximation (\S \ref{s:prelims}). We then discuss multivariate polynomial approximation (\S \ref{s:multivar-poly}), this being one of the main motivating examples for this work. The next two sections contain the core developments of this article. We describe the theory of least-squares approximation with random sampling and introduce the so-called \textit{Christoffel function}, which plays a key role in its analysis (\S \ref{s:wls-theory}). We then show that Christoffel sampling, i.e., random sampling from a probability measure whose density is proportional to the Christoffel function, leads to provably near-optimal sampling (\S \ref{s:near-opt}). The power this approach lies in its generality: Christoffel sampling is near-optimal for \textit{any} given linear approximation space. Next, we consider the matter of how much can be gained through this approach in comparison to \textit{Monte Carlo sampling}, i.e., i.i.d.\ random sampling from the problem's underlying probability measure (\S \ref{ss:MC-sampling}). Monte Carlo sampling is ubiquitous in applications, especially high-dimensional approximation tasks. Yet, as we discuss, the sample complexity of this na\"ive strategy can be arbitrarily bad. Once more, we see that the Christoffel function plays a key role in analyzing this sample complexity. Having done this, we conclude this part of the article by discussing a series of further topics (\S \ref{s:further}). In particular, we describe very recent advances of optimal (as opposed to near-optimal) sampling and its connections to sampling numbers in information-based complexity and the study of sampling discretizations in approximation theory. We also discuss connections to \textit{matrix sketching} via \textit{leverage score sampling}, as well as numerical considerations such as how to draw samples in practice.

The majority of this article considers linear approximation spaces, i.e., finite-dimensional subspaces of functions. However, modern applications increasingly make use of nonlinear spaces. Moreover, in many applications the object to recover may not be a scalar-valued function, and the samples may not be simple pointwise evaluations. In the second part, we describe a recent framework for optimal sampling with general linear samples and nonlinear approximation spaces (\S \ref{s:general-framework}). We discuss how many of the key ideas seen in linear spaces, such as Christoffel functions, naturally extend to this general setting. Finally, we end with some concluding thoughts (\S \ref{s:conclusion}).

\subsection{Scope and target audience}

In this article, we focus on foundational techniques and theory. After a brief discussion in \S \ref{s:lit-rev}, we largely omit applications.
This article is intended to be accessible to nonspecialists. We build most concepts up from first principles, relying on basic knowledge only. In order to make it as self-contained as possible, proofs of most of the results shown in this work are given in an appendix.

\section{Literature review}\label{s:lit-rev}

We commence with a short discussion of relevant literature. Additional literature on variations, extensions and further topics can be found in \S \ref{s:further}.

Least squares is a classical subject, with origins tracing back to the work of Gauss and Legendre \cite{stigler1981gauss}. Starting in the early 2010s, and motivated by problems in parametric and stochastic Differential Equations (DEs), there was a resurgence of research on this topic, focusing on high- and and infinite-dimensional function approximation, and typically involving polynomial spaces. Key works in this direction include \cite{chkifa2015discrete,cohen2013stability,migliorati2014analysis}. This resurgence was based on least squares with random sampling, inspired by Monte Carlo quadrature and its ability to integrate functions without succumbing to the \textit{curse of dimensionality}. However, it is worth noting that the goal of least-squares approximation is to achieve quasi-optimal rates of convergence with respect to the approximation space. Typically, these rates will exceed the error rate for Monte Carlo quadrature.

As noted, Monte Carlo sampling generically leads to suboptimal sample complexity bounds for least-squares approximation. This observation led to a concerted effort to develop practical sampling strategies with better performance (see \cite[\S 8.1.1]{adcock2022sparse} and \S \ref{s:further} for overviews), culminating in the near-optimal random sampling strategies which are the basis of this work. These were developed in \cite{cohen2017optimal}, but also appeared slightly earlier in \cite{hampton2015coherence} in the case of (total degree) polynomial spaces. 

At a similar time, related techniques under the name \textit{leverage score sampling}, which are based on the classical topic of \textit{statistical leverage}, have become increasingly popular in machine learning and data science. In particular, leverage score sampling is an effective tool for matrix sketching \cite{drineas2012fast,mahoney2011randomized,woodruff2014sketching}. As we comment in \S \ref{s:further}, it is can also be viewed as a special case of the techniques described in this article, corresponding to functions defined over a discrete domain. 

Finally, we mention some applications. As observed, this work is closely related to optimal design of experiments and optimal sensor placement in sampling theory and signal processing -- both large areas with countless applications that we shall not attempt to review. However, this specific line of research emerged out of computing polynomial approximations to high-dimensional functions arising in parametric and stochastic DEs \cite{berveiller2006stochastic,chkifa2015discrete,migliorati2013approximation}, and this remains a key area of application. See \cite{cohen2023near,hadigol2018least,adcock2022sparse,hampton2015coherence,guo2020constructing,narayan2015stochastic} and references therein. 
For other surveys focused multivariate polynomial approximation and parametric and stochastic DEs, see \cite{guo2020constructing,hadigol2018least} and \cite[Chpt.\ 5]{adcock2022sparse}. Note that \cite{hadigol2018least} also has an in-depth discussion on the connections to optimal design of experiments.

Recently, these techniques have also been applied to the closely related problem of numerical integration (cubature) \cite{nakatsukasa2018approximate,migliorati2022stable}.  There are also emerging applications in Trefftz methods for solving Helmoltz equations \cite{parolin2023stable} and methods for option pricing in finance \cite{ackerer2019option,filipovic2019weighted}. On the theoretical side, this line of work has also spurred recent advances in approximation theory (so-called \textit{sampling discretizations}) and information-based complexity (so-called \textit{sampling numbers}). We discuss these topics further in \S \ref{s:further}. Related ideas have also been used in sampling theory \cite{antezana2023random}. We also note that Christoffel functions are themselves useful tools for empirical inference in data analysis \cite{lasserre2022christoffel}.

Finally, through the close connection to leverage score sampling, there are various applications in machine learning and data science. These include randomized numerical linear algebra \cite{woodruff2014sketching,martinsson2020randomized}, kernel methods \cite{alaoui2015fast,avron2019universal,erdelyi2020fourier,fanuel2022nystrom,musco2017recursive} and active learning \cite{avron2019universal,chen2016statistical,derezinski2018leveraged,erdelyi2020fourier,ma2015statistical}.
Moreover, the generalization we describe in \S \ref{s:general-framework} opens the door to applications in many seemingly unrelated areas, such as inverse problems in imaging \cite{adcock2021compressive}.

\section{Preliminaries}\label{s:prelims}

Let $(D,\cD,\varrho)$ be a measure space and $L^2_{\varrho}(D)$ be the Lebesgue space of square-integrable functions $f : D \rightarrow \bbC$ with respect to $\varrho$. Typically in this work, $D \subseteq \bbR^d$.  We assume that $\varrho$ is a finite measure ($\varrho(D) < \infty$) and, therefore, without loss of generality, that $\varrho$ is a probability measure ($\varrho(D) = 1$). It is possible to consider infinite measures, but for ease of exposition we shall not do this.

Given $m \in \bbN$, we consider \textit{sampling measures} $\mu_1,\ldots,\mu_m$. These are assumed to be such that $(D,\cD,\mu_i)$ is a probability space for every $i$. We also make the following assumption.

\begin{assumption}[Absolute continuity and positivity]
\label{ass:mixture}
The additive mixture
\bes{
\mu = \frac1m \sum^{m}_{i=1} \mu_i
}
is absolutely continuous with respect to $\varrho$ and its Radon--Nikodym derivative is strictly positive almost everywhere on $\mathrm{supp}(\varrho)$.
\end{assumption}

Often, we assume that $\mu_1 = \cdots = \mu_m = \mu$. But, as we shall see later, the flexibility of allowing a different sampling measure for each sample point is convenient in some settings.
Assumption \ref{ass:mixture} allows us to write
\be{
\label{mu_weight_fn}
\frac1m \sum^{m}_{i=1} \D \mu_i(x) = \nu(x) \D \varrho(x),
} 
where the density $\nu : D \rightarrow \bbR$ (the Radon--Nikodym derivative) is measurable, positive almost everywhere and satisfies
\be{
\label{w_normalization}
\int_D \nu(x) \D \varrho(x) = 1.
}
In what follows it will often be more convenient to work with the reciprocal of this function. We define the \textit{weight function} $w : D \rightarrow \bbR$ as $w(x) = 1/\nu(x)$, $x \in D$.

Given sampling measures $\mu_1,\ldots,\mu_m$, we now draw samples $x_i \sim \mu_i$, $i = 1,\ldots,m$, independently and consider noisy measurements of an unknown function $f : D \rightarrow \bbC$ of the form  
\be{
\label{f_meas}
y_i = f(x_i) + e_i ,\quad i = 1,\ldots,m.
}
Typically, we will assume that $f \in L^2_{\varrho}(D)$ so that the samples \ef{f_meas} are almost surely well defined.

We consider a \textit{bounded, deterministic} noise model, where the $e_i$'s are not random, but instead are assumed to be small in magnitude. Our aim is to derive error bounds in which the noise term depends linearly on
\bes{
\frac{\nms{e}_2}{\sqrt{m}} = \sqrt{\frac1m \sum^{m}_{i=1} |e_i|^2},\qquad \text{where } e = (e_i)^{m}_{i=1}.
}
Notice that $\nm{e}_2 / \sqrt{m} \leq \nm{e}_{\infty} = \max_{i=1,\ldots,m} | e_i|$. Hence such error bounds allow for a constant amount of corruption in each sample $y_i$.
Random noise models (including unbounded noise)  can also be considered (see \cite{migliorati2015convergence,cohen2017optimal} and \cite[Rem.\ 5.1]{adcock2022sparse}).

\begin{table}[t]
\begin{longtable}{@{} p{.25\textwidth}  p{.70\textwidth} } 
\hline
$(D,\cD,\varrho)$ & Measure space
\\
$L^p_{\varrho}(D)$ & Lebesgue space of functions $D \rightarrow \bbC$, where $1 \leq p \leq \infty$
\\
$\ip{\cdot}{\cdot}_{L^p_{\varrho}(D)}$ & Inner product on $L^2_{\varrho}(D)$
\\
$\nms{\cdot}_{L^p_{\varrho}(D)}$ & Norm on $L^p_{\varrho}(D)$
\\
$f : D \rightarrow \bbC$ & Function to approximate
\\
$\mu_1,\ldots,\mu_m$ & Sampling measures
\\
$w$ & Weight function, given by \ef{mu_weight_fn}
\\
$x_1,\ldots,x_m$ & Sample points, where $x_i \sim \mu_i$ independently for $i = 1,\ldots,m$
\\
$y_i = f(x_i) + e_i$ & Noisy samples of $f$
\\
 $\cP \subseteq L^2_{\varrho}(D)$ & Finite-dimensional subspace in which to approximate $f$
 \\
 $n$ & Dimension of $\cP$
 \\
 $A$, $b$ & Matrix and measurement vector of the algebraic least-squares problem, given by \ef{ls-Ab}
 \\
 $\ip{\cdot}{\cdot}_{\mathsf{disc},w}$, $\nms{\cdot}_{\mathsf{disc},w}$ & Semi-inner product and seminorm defined by the sample points, given by \ef{semi-inner-product} and \ef{seminorm}, respectively
\\
\hline
\end{longtable}
\caption{A summary of the main notation used in this paper.}
\label{tab:summary}
\end{table}

\subsection{Weighted least-squares approximation}

Let $\cP \subseteq L^2_{\varrho}(D)$ be an arbitrary $n$-dimensional subspace, where $n \leq m$, in which we seek to approximate the unknown $f$ using the measurements \ef{f_meas}. We term $\cP$ the \textit{approximation space}. We consider general approximation spaces in this work. In particular, this means that \textit{interpolation} is generally impossible \cite{dolbeault2021optimal}, as this requires intricate constructions of sample points that rely heavily on the structure of $\cP$. Instead, we consider the \textit{weighted least-squares approximation}
\be{
\label{wls-prob}
\hat{f} \in \argmin{p \in \cP} \frac1m \sum^{m}_{i=1} w(x_i) | y_i - p(x_i) |^2.
}
Note that the loss function is almost surely well defined for any fixed $f \in L^2_{\varrho}(D)$ and weight function $w$ as above, and any $p \in \cP$ (since $\cP$ is finite dimensional).

\rem{
[Scaling factors]
The scaling factors in \ef{wls-prob} are motivated by noticing that
\be{
\label{exp-sum-scaling}
\bbE \left [ \frac1m \sum^{m}_{i=1} w(x_i) | g(x_i) |^2 \right ]= \frac1m \sum^{m}_{i=1} \int_{D} w(x) |g(x)|^2 \D \mu_i(x)
 =  \int_{D} |g(x)|^2 \D \rho(x) = \nms{g}^2_{L^2_{\varrho}(D)},
}
where the second equality is due to \ef{mu_weight_fn}. Thus, in the noiseless case, \ef{wls-prob} can be considered as a empirical approximation to the \textit{continuous} least-squares approximation
\be{
\label{cts-min}
\hat{f} = \argmin{p \in \cP } \nms{f - p}^2_{L^2_{\varrho}(D)},
}
i.e., the best approximation to $f$ from $\cP$ in the $L^2_{\varrho}(D)$-norm. In particular, if $\mu_1 = \cdots = \mu_m = \mu$, then the minimizers of \ef{wls-prob} converge almost surely to the minimizer of \ef{cts-min} as $m \rightarrow \infty$ \cite{guo2020constructing}.
}

The objective of this article is to describe how to choose the measures $\mu_1,\ldots,\mu_m$ to achieve the most sample-efficient approximation. We shall compare such strategies against the standard approach of \textit{Monte Carlo sampling}, i.e., i.i.d.\ random sampling from $\varrho$. This is equivalent to setting
\bes{
\mu_1 = \cdots = \mu_m = \varrho,
}
which leads, via \ef{mu_weight_fn}, to $\nu  \equiv 1$. In this case, \ef{wls-prob} is an \textit{unweighted} least-squares approximation.

\rem{
[Hierarchical approximation]
\label{rem:hierarchical}
Often, rather than a fixed subspace $\cP$, one may wish to construct a sequence of approximations in a \textit{nested} collection of subspaces
\bes{
\cP^{(1)} \subseteq \cP^{(2)} \subseteq \cdots,
}
of dimension $\dim(\cP^{(k)}) = n_k$. Given integers $1 \leq m_1 \leq m_2 \leq \cdots$ satisfying $m_k \geq n_k$, $\forall k$, one then aims to design a nested collection of sample points
\bes{
\{ x^{(1)}_i \}^{m_1}_{i=1} \subseteq \{ x^{(2)}_{i} \}^{m_2}_{i=1} \subseteq \cdots.
}
We write $\hat{f}^{(1)},\hat{f}^{(2)},\ldots$ for the ensuing (weighted) least-squares approximations, where $\hat{f}^{(k)}$ is constructed from the sample points $\{ x^{(k)}_i \}^{m_k}_{i=1}$. Nestedness implies that samples are recycled at each iteration -- a highly desirable property in the setting of limited data.
We call such a procedure a \textit{hierarchical} approximation scheme (the terms \textit{progressive} \cite{cohen2015approximation} or \textit{sequential} \cite{arras2019sequential} are also used).
}

\subsection{Reformulations of \ef{wls-prob}}

Given a basis $\{ \phi_i \}^{n}_{i=1}$ for $\cP$, the problem \ef{wls-prob} is easily reformulated as an algebraic least-squares problem for the \textit{coefficients} $\hat{c} = (\hat{c}_i)^{n}_{i=1} \in \bbC^n$ of $\hat{f} = \sum^{n}_{i=1} \hat{c}_i \phi_i$. This takes the form
\be{
\label{algls-prob}
\hat{c} \in \argmin{c \in \bbC^n} \nms{A c - b}^2_{2},
}
where
\be{
\label{ls-Ab}
A = \left ( \sqrt{\frac{w(x_i)}{m}} \phi_j(x_i) \right )^{m,n}_{i,j=1} \in \bbC^{m \times n},\qquad b = \left ( \sqrt{\frac{w(x_i)}{m}} (f(x_i) + e_i ) \right )^{m}_{i=1} \in \bbC^m.
}
To be precise, every minimizer $\hat{f}$ satisfying \ef{wls-prob} has coefficients $\hat{c}$ that satisfy \ef{algls-prob} and vice versa. Classical least-squares analysis asserts that any vector $\hat{c}$ satisfying \ef{algls-prob} is also a solution of the \textit{normal equations}
\be{
\label{normal-eqns}
A^* A c = A^* b
}
and vice versa. Rewriting the normal equations in terms of functions also leads to the following variational form of \ef{wls-prob}:
\be{
\label{variational-form}
\text{Find $\hat{f} \in \cP$ such that }\ip{\hat{f}}{p}_{\mathsf{disc},w} = \ip{f}{p}_{\mathsf{disc},w} + \frac1m \sum^{m}_{i=1} w(x_i) e_i \overline{p(x_i)},\ \forall p \in \cP.
}
This is equivalent to \ef{normal-eqns} in the same sense as before. Here we wrote
\be{
\label{semi-inner-product}
\ip{g}{h}_{\mathsf{disc},w} = \frac1m \sum^{m}_{i=1} w(x_i) \overline{g(x_i)} h(x_i),
}
for the discrete semi-inner product induced by the sample points and the weight function (whenever defined). For convenience, we shall denote the corresponding seminorm as
\be{
\label{seminorm}
\nm{g}_{\mathsf{disc},w} = \sqrt{\frac1m \sum^{m}_{i=1} w(x_i) | g(x_i) |^2}.
}
In the noiseless case $e =0$, \ef{variational-form} implies that $\hat{f}$ is precisely the orthogonal projection of $f$ onto $\cP$ with respect to the discrete semi-inner product \ef{semi-inner-product}. Since \ef{semi-inner-product} is an empirical approximation to the continuous inner product  $\ip{\cdot}{\cdot}_{L^{2}_{\varrho}(D)}$ (recall \ef{exp-sum-scaling}), this sheds further light on why minimizers of \ef{wls-prob} generally converge to \ef{cts-min} (the orthogonal projection in the $L^2_{\varrho}$-inner product).

\rem{[Numerical considerations]
\label{rem:numerical}
Fast numerical computations are not the primary concern of this article. However, we note in passing that \ef{algls-prob} can be solved using standard linear algebra techniques. Since the matrix $A$ is generally dense and unstructured, each matrix-vector multiplication involves $\ord{m n}$ \textit{floating-point operations (flops)}. Hence, when using an iterative method such as conjugate gradients, the number of flops that suffice to compute $\hat{c}$ to an error of $\eta > 0$ (in the norm $\nm{A \cdot }_2$ ) is roughly $\mathrm{cond}(A) \cdot m \cdot n \cdot \log(1/\eta)$, where $\mathrm{cond}(A)$ is the condition number of $A$. In \S \ref{s:wls-theory} we see that the sufficient conditions that ensure accuracy and stability of the approximation $\hat{f}$ also guarantee that $A$ is well conditioned.
}

\subsection{Key terminology}\label{ss:terminology}

We conclude this section by introducing some key terminology that will be used from now on. 
First, we say that the approximation $\hat{f}$ is \textit{$L^{2}_{\varrho}$-quasi-optimal} or \textit{$(L^{2}_{\varrho},L^{\infty}_{\varrho})$-quasi-optimal} if, in the absence of noise,
\bes{
\nm{f - \hat{f}}_{L^2_{\varrho}(D)} \lesssim \inf_{p \in \cP} \nm{f - p}_{L^2_{\varrho}(D)}\quad \text{or}\quad 
\nm{f - \hat{f}}_{L^2_{\varrho}(D)} \lesssim \inf_{p \in \cP} \nm{f - p}_{L^{\infty}_{\varrho}(D)},
}
respectively (note that the term \textit{instance optimality} is also used \cite{dolbeault2021optimal}). Here and elsewhere, we write $a \lesssim b$ to mean that the is a numerical constant $C>0$ such that $a \leq C b$. We also write $a \asymp b$ whenever $a \lesssim b$ and $b \lesssim a$.
Quasi-optimality implies that the error of $\hat{f}$ is proportional to the \textit{best approximation} error of $f$ from $\cP$, measured in some norm.
Since $L^{2}_{\varrho}$-quasi-optimality is stronger than $(L^{2}_{\varrho},L^{\infty}_{\varrho})$-quasi-optimality, achieving it will be our main goal.  Moving to the matter of noise, we say that the recovery is \textit{stable} if, given noisy samples, the recovery error scales linearly in $\nm{e}_2 / \sqrt{m}$, i.e.,
\bes{
\nm{f - \hat{f}}_{L^2_{\varrho}(D)} \lesssim e_{\cP}(f) + \nm{e}_{2} /\sqrt{m},
}
where $e_{\cP}(f)$ is some best approximation error term. Finally, we say that a sampling strategy (i.e., a collection of measures $\mu_1,\ldots,\mu_m$) has \textit{near-optimal sample complexity} or \textit{optimal sample complexity} if a quasi-optimal and stable approximation is obtained whenever $m \geq C n \log(n)$ or $m \geq C n$, respectively, for some constant $C > 0$.

\section{Application to multivariate polynomial approximation}\label{s:multivar-poly}

We now introduce an important example considered in this paper, namely, multivariate polynomial approximation in $d \geq 1$ dimensions.

\subsection{Spaces of multivariate polynomials}\label{ss:poly-spaces}

Let $D \subseteq \bbR^d$, $\bbN_0 = \{0,1,2\ldots\}$ be the set of nonnegative integers and $S \subset \bbN^d_0$ be a finite set of multi-indices with $|S| = n$, and consider the polynomial space
\be{
\label{PS-def}
\cP = \bbP_{S}  : = \spn \left \{ x \mapsto x^{\nu} : \nu \in S \right \} \subset L^2_{\varrho}(D).
}
Here, $x \in D$ denotes the $d$-dimensional variable, $\nu \in \bbN^d_0$ is a multi-index and $x^{\nu}$ denotes the mult-index power. In other words, if $x = (x_1,\ldots,x_d)$ and $\nu = (\nu_1,\ldots,\nu_d)$ then $x^{\nu} = x^{\nu_1}_{1} \cdots x^{\nu_d}_{d}$.
There are several standard choices for the index set $S$. In low dimensions, it is common to use the (isotropic) \textit{tensor-product} or \textit{total degree} index sets of \textit{order} $p \in \bbN_0$. These are given by
\be{
\label{TP-TD}
S = S^{\mathsf{TP}}_p = \left \{ \nu \in \bbN^d_0 : \max_{k=1,\ldots,d} \nu_k \leq p  \right \},\qquad
S = S^{\mathsf{TD}}_p = \left \{ \nu \in \bbN^d_0 : \sum^{d}_{k=1} \nu_k \leq p \right \}
}
respectively. Unfortunately, the cardinalities of these index sets grow rapidly in $d$ (for fixed $p$). Indeed, $S^{\mathsf{TP}}_p = (p+1)^d \sim p^d$ and $|S^{\mathsf{TD}}_p | = {p+d \choose d} \sim \frac{d^p}{p!}$ as $d \rightarrow \infty$ with $p$ fixed. A better choice in moderate dimensions is the \textit{hyperbolic cross} index set
\be{
\label{HC}
S = S^{\mathsf{HC}}_p  =  \left \{ \nu \in \bbN^d_0 : \prod^{d}_{k=1} (\nu_k+1) \leq p +1 \right \}.
}
However, as $d$ increases, this may also become too large to use. Since high-dimensional functions often have a very anisotropic dependence with respect to the coordinate variables, when $d$ is large one may also consider anisotropic versions of these index sets. Given an anisotropy parameter $a = (a_k)^{d}_{k=1}$ with $a > 0$ (understood componentwise) and $p \geq 0$ (not necessarily an integer), the corresponding anisotropic index sets are defined as
\eas{
S^{\mathsf{TP}}_{p,a} = \left \{ \nu \in \bbN^d_0 : \max_{k=1,\ldots,d} a_k \nu_k \leq p  \right \},
\qquad
S^{\mathsf{TD}}_{p,a} = \left \{ \nu \in \bbN^d_0 : \sum^{d}_{k=1} a_k \nu_k \leq p \right \}
}
and
\bes{
S^{\mathsf{HC}}_{p,a}  =  \left \{ \nu \in \bbN^d_0 : \prod^{d}_{k=1} (\nu_k+1)^{a_k} \leq p +1 \right \}.
}
Notice that the isotropic index sets are recovered by setting $a = 1$ (the vector of ones).

The choice of index set is not the focus of this paper. We remark, however, that all index sets defined above are examples of \textit{lower} (also known as \textit{monotone} or \textit{downward closed}) sets. A set $S \subseteq \bbN^d_0$ is \textit{lower} if whenever $\nu \in S$ and $\mu \leq \nu$ (understood componentwise, once more), one has that $\mu \in S$. Lower sets are often desirable in multivariate polynomial approximation. See \cite{cohen2015approximation,guo2020constructing}, \cite[Chpts.\ 2 \& 5]{adcock2022sparse} and references therein for further discussion.

\subsection{Multivariate orthogonal polynomials on tensor-product domains}\label{ss:orth-polys}

As we shall see in the next section, orthonormal bases play a key role in least-squares approximation from random samples. It is therefore convenient have an explicit orthonormal basis for $\bbP_S$.
When $S$ is lower and $D$ and $\varrho$ are of tensor-product type, such a basis is easily generated via taking tensor products of univariate orthogonal polynomials. For concreteness, let
\bes{
D = (a_1,b_1) \times \cdots \times (a_d , b_d),\qquad \varrho = \rho_1 \times \cdots \times \rho_d,
}
where, for each $k = 1,\ldots,d$,  $-\infty \leq a_k < b_k \leq \infty$ and $\rho_k$ is a probability measure on $(a_k,b_d)$. Then, under mild conditions on $\rho_k$ (see, e.g., \cite[\S 2.2]{szego1975orthogonal}), there exists a unique sequence of orthonormal polynomials
\bes{
\{ \psi^{(k)}_i \}^{\infty}_{i=0} \subset L^2_{\rho_k}(a_k,b_k),
}
where, for each $i$, $\psi^{(k)}_i$ is a polynomial of degree $i$. 
Using this, one immediately obtains an orthonormal basis of $L^2_{\varrho}(D)$ via tensor products. Specifically,
\bes{
\{ \Psi_{\nu} \}_{\nu \in \bbN^d_0} \subset L^2_{\varrho}(D),\quad \text{where }\Psi_{\nu} = \psi^{(1)}_{\nu_1} \otimes \cdots \otimes \psi^{(d)}_{\nu_d},\ \forall \nu = (\nu_k)^{d}_{k=1} \in \bbN^d_0.
}
What about the subspace $\bbP_S$ introduced in \ef{PS-def}? Fortunately, when $S$ is a lower set the functions $\Psi_{\nu}$ with indices $\nu \in S$ also form an orthonormal basis for this space. Namely,
\bes{
\text{$S$ lower}\ \Longrightarrow\ \spn \{ \Psi_{\nu} : \nu \in S \} = \bbP_S.
}
See, e.g., \cite[Proof of Thm.\ 6.5]{adcock2020approximating}.
This property, combined with the tensor-product structure of the basis functions, makes the sampling strategies we devise later in \S \ref{s:near-opt} computationally feasible for the space $\bbP_S$. We discuss such computational considerations further in \S \ref{ss:practical-sampling}.

To conclude this section, we now list several standard families of univariate measures and their corresponding orthogonal polynomials \cite{gautschi2004orthogonal,szego1975orthogonal}. Consider a compact interval, which without loss of generality we take to be $(-1,1)$. For $\alpha,\beta > -1$, the Jacobi (probability) measure is given by
\bes{
\D \rho(x) = C_{\alpha,\beta} (1-x)^{\alpha} (1+x)^{\beta} \D x,\qquad \text{where }C_{\alpha,\beta} = \left ( \int^{1}_{-1} (1-x)^{\alpha} (1+x)^{\beta} \D x \right )^{-1}.
}
This measure generates the \textit{Jacobi} polynomials for general $\alpha,\beta$ and the \textit{ultraspherical} polynomials when $\alpha = \beta$. Of particular interest are the following cases.
\bullc{
\item $\alpha = \beta = -1/2$. This corresponds to the arcsine measure $\D \rho(x) = ( \pi \sqrt{1-x^2})^{-1} \D x$ and yields the \textit{Chebyshev} polynomials of the first kind.
\item $\alpha = \beta = 0$. This corresponds to the uniform measure $\D \rho(x) = \frac12 \D x$ and yields the \textit{Legendre} polynomials. 
\item $\alpha = \beta = 1/2$. This corresponds to the measure $\D \rho(x) = (2 / \pi) \sqrt{1-x^2} \D x$ and yields the \textit{Chebyshev} polynomials of the second kind.
}
We will consider these polynomials later in this paper.
We will also briefly discuss certain unbounded domains. Here two common examples are
\bullc{
\item $\D \rho(x) = (2 \pi)^{-1/2} \E^{-x^2/2} \D x$ over $\bbR$, which yields the \textit{Hermite} polynomials,
\item $\D \rho(x) = \E^{-x} \D x$ over $[0,\infty)$, which yields the \textit{Laguerre} polynomials.
}

\section{Theory of weighted least-squares approximation from random samples}\label{s:wls-theory}

We now return to the general setting, where $\cP \subseteq L^2_{\varrho}(D)$ is an arbitrary $n$-dimensional subspace. In this section, we present a series of results on the accuracy, stability and sample complexity of weighted least-squares approximation from random samples.

\subsection{Basic accuracy and stability guarantee}

Accuracy and stability of weighted least-squares approximation are controlled by the following \textit{discrete stability constants}:
\eas{
\alpha_{w} = \inf \left \{ \nm{p}_{\mathsf{disc},w} : p \in \cP,\ \nms{p}_{L^2_{\varrho}(D)} = 1 \right \},
\quad
 \beta_w =  \sup \left \{ \nm{p}_{\mathsf{disc},w} : p \in \cP,\ \nms{p}_{L^2_{\varrho}(D)} = 1 \right \}.
}
Note that if $\{ \phi_i \}^{n}_{i=1}$ is an orthonormal basis for $\cP$, then it is straightforward to show that
\be{
\label{alpha-beta-sigma}
\alpha_{w} = \sigma_{\min}(A) = \sqrt{\lambda_{\min}(A^*A)},\quad \beta_{w} = \sigma_{\max}(A) = \sqrt{\lambda_{\max}(A^*A)},
}
where $A$ is the least-squares matrix \eqref{ls-Ab}. 

By definition, $\alpha_w$ and $\beta_w$ are the optimal constants in the inequalities
\be{
\label{MZ-inequality}
\alpha_w \nms{p}_{L^2_{\varrho}(D)} \leq \nm{p}_{\mathsf{disc},w} \leq \beta_w \nms{p}_{L^2_{\varrho}(D)},\quad \forall p \in \cP.
}
Hence, if $0 < \alpha_w \leq \beta_w < \infty$ then $\nms{\cdot}_{\mathsf{disc},w}$ is a norm over the  subspace $\cP$, with $\alpha_w,\beta_w$ being the constants of the equivalence between it and the $L^2_{\varrho}$-norm. Squaring and writing out the discrete seminorm, we see that \ef{MZ-inequality} is equivalent to
\be{
\label{samp-disc}
\alpha^2_w \nms{p}^2_{L^2_{\varrho}(D)} \leq \frac1m \sum^{m}_{i=1} w(x_i) | p(x_i) |^2 \leq \beta^2_w \nms{p}^2_{L^2_{\varrho}(D)},\quad \forall p \in \cP.
}
In approximation theory, this is known as a \textit{sampling discretization} \cite{kashin2022sampling} or a (weighted) \textit{Marcinkiewicz--Zygmund} inequality \cite{temlyakov2018marcinkiewicz}.

\lem{
[Accuracy and stability of weighted least squares]
\label{l:wLS-acc-stab}
Let $\cP \subset L^2_{\varrho}(D)$, $f \in  L^2_{\varrho}(D)$, $x_1,\ldots,x_m \in D$ be sample points at which both $f$ and any $p \in \cP$ are defined, $e \in \bbC^m$ and $w : D \rightarrow \bbR$ be such that $w(x_i) > 0$, $\forall i \in \{1,\ldots,m\}$. Suppose that $\alpha_w > 0$. Then the weighted least-squares problem \ef{wls-prob} has a unique solution $\hat{f}$. Moreover, this solution satisfies
\be{
\label{LSerrbd}
\nm{f - \hat{f}}_{L^2_{\varrho}(D)} \leq \inf_{p \in \cP} \left \{ \nm{f - p}_{L^{2}_{\varrho}(D)} + \frac{1}{\alpha_w} \nm{f - p}_{\mathsf{disc},w} \right \} + \frac{1}{\alpha_w} \nms{e}_{2,w},
}
where $\nms{e}_{2,w} = \sqrt{\frac1m \sum^{m}_{i=1} w(x_i) | e_i |^2 } $.
Also, if $\{ \phi_i \}^{n}_{i=1}$ is an orthonormal basis of $\cP$ then the condition number of the least-squares matrix \ef{ls-Ab} satisfies $\mathrm{cond}(A) = \beta_w / \alpha_w$.
}

The condition number statement in this result follows immediately from \ef{alpha-beta-sigma}. The other parts are a standard exercise -- see the appendix for a short proof. Observe that this result holds for arbitrary weight functions $w$ and sample points $x_1,\ldots,x_m$ satisfying the stipulated assumptions. At this stage, we do not require the sample points to be random. This will be used in the next subsection to derive concrete sample complexity estimates.

\rem{
[The noise bound]
\label{rem:noise-bound}
On the face of it, the noise term $\nms{e}_{2,w}$ is undesirable since terms $e_i$ corresponding to large values of $w(x_i)$ are more heavily weighted than others. We will take this into account later when we construct near-optimal sampling measures. Specifically, in \S \ref{ss:near-opt-samp} we construct sampling measures that lead to log-linear sample complexity and for which $w(x) \leq 2$. Hence, the noise term $\nms{e}_{2,w} \leq \sqrt{2} \nms{e}_{2} / \sqrt{m}$ in this case, which implies that the weighted least-squares approximation is stable in the sense of \S \ref{ss:terminology}.
}

\subsection{The (reciprocal) Christoffel function}

We now consider the main setting of this paper, where the samples points are drawn randomly and independently with $x_i \sim \mu_i$, $i=1,\ldots,m$, for measures $\mu_i$ satisfying Assumption \ref{ass:mixture}. Our aim is to analyze the sample complexity of weighted least-squares approximation. In view of Lemma \ref{l:wLS-acc-stab}, this involves first bounding the lower discrete stability constants $\alpha_w$ and $\beta_w$.

A key tool in this analysis is the \textit{Christoffel function} of $\cP$. Christoffel functions are well-known objects in approximation theory \cite{nevai1986geza,xu1995christoffel}, where they are typically considered in the context of spaces spanned by algebraic polynomials. It transpires that Christoffel functions -- or, more precisely, their reciprocals -- are also fundamentally associated with random sampling for least-squares approximation.

\defn{[Christoffel function]
\label{def:christoffel}
Let $\cP \subseteq L^2_{\varrho}(D)$. The \textit{(reciprocal) Christoffel function} of $\cP$ is the function $\cK = \cK(\cP) : D \rightarrow \bbR$ defined by
\be{
\label{Kappa-def}
\cK(x) = \cK(\cP)(x) : = \sup \left \{ \frac{| p(x) |^2}{\nm{p}^2_{L^2_{\varrho}(D)}} : p \in \cP,\ p \neq 0 \right \}.
}
}
In other words, $\cK(x)$ measures how large in magnitude an element of $\cP$ can be at $x \in D$ in relation to its $L^2_{\varrho}$-norm.
This function also admits an explicit expression. Given an arbitrary orthonormal basis $\{ \phi_i \}^{n}_{i=1}$ of $\cP$, it is a short exercise to show that
\be{
\label{Kappa-def-alt}
\cK(x) = \sum^{n}_{i=1} | \phi_i(x) |^2.
}
Often taken as the definition of $\cK$, this formulation is particularly useful in subsequent analysis. It also emphasizes the fact that $\cK$ is precisely the diagonal of the \textit{Christoffel--Darboux} kernel associated with $\cP$, i.e., the reproducing kernel of $\cP$ in $L^2_{\varrho}(D)$ \cite[\S 3]{nevai1986geza}. 

For reasons that will become clear soon, we are particularly interested in the maximal behaviour of the function $w(x) \cK(x)$, where $w = 1/\nu$ for $\nu$ defined by \ef{mu_weight_fn}. We therefore let
\be{
\label{kappa-w-def}
\kappa_w = \kappa_w(\cP) : = \esssup_{x \sim \varrho} w(x) \cK(\cP)(x).
}
To continue the connection with approximation theory, it is worth noting that $\kappa_w$ is the optimal constant in the (weighted) \textit{Nikolskii-type inequality} (see, e.g., \cite{milovanovic1994topics} and references therein),
\be{
\label{nikolskii}
\nm{\sqrt{w(\cdot)} p(\cdot) }_{L^{\infty}_{\varrho}(D)} \leq \sqrt{\kappa_w} \nm{p}_{L^2_{\varrho}(D)},\quad \forall p \in \cP.
}
Thus, $\kappa_w$ measures how large the scaled element $\sqrt{w(\cdot)} p(\cdot)$ can be uniformly in relation to the $L^2_{\varrho}$-norm of $p$.
It is important to observe that
\be{
\label{kappa-w-lb}
\kappa_w(\cP) \geq n,
}
for any weight function $w$ and $n$-dimensional subspace $\cP$. This bound follows from the observation that $\int_{D} \cK(x) \D \varrho(x) = n$, which is itself an immediate consequence of \ef{Kappa-def-alt}. Using this, \ef{w_normalization} and the fact that $w = 1/\nu$, we get 
\bes{
n = \int_{D} \cK(x) \D \varrho(x) =  \int_{D} w(x) \cK(x) \frac{1}{w(x)} \D \varrho(x) \leq \kappa_w \int_{D} \frac{1}{w(x)} \D \varrho(x) = \kappa_w,
}
as required.

\subsection{Bounding the discrete stability constants}

The following result establishes a key relationship between the Christoffel function and the sample complexity of weighted least-squares approximation with random samples.

\thm{
[Estimates for $\alpha_w$ and $\beta_w$ in probability]
\label{t:alpha-beta-est}
Let $0 < \delta,\epsilon < 1$, $\cP \subset L^2_{\varrho}(D)$ be a finite-dimensional subspace with $\dim(\cP) = n$ and $\mu_1,\ldots,\mu_m$ be probability measures satisfying Assumption \ref{ass:mixture}. Consider sample points drawn randomly and independently with $x_i \sim \mu_i$, $i = 1,\ldots,m$. Then
\be{
\label{alpha-beta-delta}
\sqrt{1-\delta} < \alpha_w \leq \beta_w < \sqrt{1+\delta}
}
with probability at least $1-\epsilon$, provided
\be{
\label{m-bound-alpha-beta}
m \geq C_{\delta} \cdot \kappa_w(\cP) \cdot \log(2 n/\epsilon),\quad \text{where }C_{\delta} = ((1+\delta) \log(1+\delta) - \delta))^{-1}
}
and $w = 1/\nu$ and $\kappa_w$ are as in \ef{mu_weight_fn} and \ef{kappa-w-def}, respectively.
}

This result is well known. In view of \ef{alpha-beta-sigma}, its proof relies on bounding the maximum and minimum eigenvalues of $A^*A$. This is achieved by using what have now become quite standard \textit{matrix concentration} inequalities, such as the \textit{matrix Chernoff} bound \cite[Thm.\ 1.1]{tropp2012user-friendly}. This bound is presented as Theorem \ref{t:matrix-chernoff} in the appendix, along with the proof of Theorem \ref{t:alpha-beta-est}.

\rem{
[One-sided estimates]
The conclusions of Lemma \ref{l:wLS-acc-stab} only rely on bounding the lower discrete stability constant $\alpha_w$ from below. This can be done with a slightly smaller sampling condition than \ef{m-bound-alpha-beta}. It follows readily from the proof of Theorem \ref{t:alpha-beta-est} that $\alpha_w > \sqrt{1-\delta}$
with probability at least $1-\epsilon$, whenever
\bes{
m \geq C'_{\delta} \cdot \kappa_w(\cP) \cdot \log(n/\epsilon),\quad \text{where }C'_{\delta} = ((1-\delta) \log(1-\delta) + \delta)^{-1}.
}
However, bounding $\beta_w$ from above yields an upper bound on the condition number of $A$ (see Lemma \ref{l:wLS-acc-stab}), which, as discussed in Remark \ref{rem:numerical}, is important for numerical purposes.
}

\subsection{Error bounds in probability}\label{ss:err-bds-prob}

We next combine Lemma \ref{l:wLS-acc-stab} and Theorem \ref{t:alpha-beta-est} to obtain error bounds for weighted least-squares approximation. We split these bounds into two types: error bounds in probability (this subsection) and error bounds in expectation (the next subsection). In these two subsections, we will strive for generality by tracking the dependence in these bounds on the parameter $0 < \delta < 1$ appearing in Theorem \ref{t:alpha-beta-est}. However, it is informative to think of this as a fixed scalar, e.g., $\delta = 1/2$.

\cor{
[First uniform error bound in probability]
\label{cor:err-prob-unif-1}
Let $0 < \delta,\epsilon < 1$, $\cP \subset L^2_{\varrho}(D)$ be a finite-dimensional subspace with $\dim(\cP) = n$ and $\mu_1,\ldots,\mu_m$ be probability measures satisfying Assumption \ref{ass:mixture}. Consider sample points drawn randomly and independently with $x_i \sim \mu_i$, $i = 1,\ldots,m$, where 
\bes{
m \geq C_{\delta} \cdot \kappa_w(\cP) \cdot \log(2 n /\epsilon),\qquad C_{\delta} = ((1+\delta) \log(1+\delta) - \delta))^{-1}
}
and $w= 1/\nu$ and $\kappa_w$ are as in \ef{mu_weight_fn} and \ef{kappa-w-def}, respectively. Then the following hold with probability at least $1-\epsilon$. For any $e \in \bbC^m$ and $f \in L^2_{\varrho}(D)$ that is defined everywhere in $D$, the weighted least-squares approximation $\hat{f}$ is unique and satisfies
\be{
\label{ls-err-bd-prob-unif-1}
\nm{f - \hat{f}}_{L^2_{\varrho}(D)} \leq \left ( 1 +  \frac{1}{\sqrt{1-\delta}} C_w \right )  \inf_{p \in \cP} \nm{f - p}_{L^{\infty}_{\varrho}(D)} + \frac{1}{\sqrt{1-\delta}} \nms{e}_{2,w},
}
where $C_w = \sqrt{\frac1m \sum^{m}_{i=1} w(x_i)}$. Moreover, if $\{ \phi_i \}^{n}_{i=1}$ is an orthonormal basis of $\cP$, then the condition number of the least-squares matrix \ef{ls-Ab} satisfies $\mathrm{cond}(A) \leq \sqrt{\frac{1+\delta}{1-\delta}}$.
}

This result follows immediately from Lemma \ref{l:wLS-acc-stab} and Theorem \ref{t:alpha-beta-est} via the estimate
\bes{
\nms{f-p}_{\mathsf{disc},w} \leq C_w \nm{f - p}_{L^{\infty}_{\varrho}(D)}.
}
Now suppose that $\delta = 1/2$ (for concreteness) and assume further that $w(x) \lesssim 1$, a.e.\ $x \sim \varrho$. This will be the case in \S \ref{s:near-opt} when we construct near-optimal sampling measures. Then \ef{ls-err-bd-prob-unif-1} reads
\bes{
\nm{f - \hat{f}}_{L^2_{\varrho}(D)} \lesssim \inf_{p \in \cP} \nm{f - p}_{L^{\infty}_{\varrho}(D)} + \nms{e}_{2} / \sqrt{m}.
}
Using the terminology introduced in \S \ref{ss:terminology}, we conclude that the approximation is stable and $(L^2_{\varrho},L^{\infty}_{\varrho})$-quasi-optimal.
In some problems, the difference between this and $L^{2}_{\varrho}$-quasi-optimality may be of little consequence. For example, in the case of polynomial approximation of holomorphic functions in low dimensions, the best approximation error decays exponentially fast with respect to $n$ in both the $L^{\infty}_{\varrho}$- and $L^2_{\varrho}$-norms (see, e.g., \cite[\S 3.5-3.6]{adcock2022sparse}). On the other hand, for high-dimensional holomorphic functions or functions of finite regularity in any dimension, the best approximation errors decay algebraically fast, with, typically, the $L^{\infty}_{\varrho}$-norm error decaying at least $\ord{\sqrt{n}}$ slower than the $L^{2}_{\varrho}$-norm error (see, e.g., \cite[\S 3.8-3.9]{adcock2022sparse}). Thus, the crude bound \ef{ls-err-bd-prob-unif-1} may underestimate the convergence rate of the least-squares approximation. Motivated by these considerations, we next discuss how to establish $L^2_{\varrho}$-quasi-optimality results.

\rem{[Uniform versus nonuniform]
\label{rem:unif-nonunif}
Corollary \ref{cor:err-prob-unif-1} is a \textit{uniform} result, in the sense that a single random draw of the sample points suffices for \textit{all} functions. We next discuss nonuniform results, in which the error bound holds with high probability for each fixed function. Uniform bounds are desirable in many applications, as it means that the same sample points (which may correspond to, e.g., sensor locations) can be re-used for approximating multiple functions. They also allow one to derive \textit{worst-case} error bounds. Indeed, let $\cF \subset L^2_{\varrho}(D)$ be a set of functions that are defined everywhere and for which
\bes{
E_{\cP}(\cF) = \sup_{f \in \cF} \inf_{p \in \cP} \nm{f - p}_{L^{\infty}_{\varrho}(D)} < \infty.
}
Often $\cF$ is a unit ball of some Banach space -- for example, the Sobolev space $H^k_{\varrho}(D)$. Then, ignoring noise for simplicity and assuming as before that $\delta = 1/2$ and $w(x) \lesssim 1$, a.e.\ $x \sim \varrho$, Corollary \ref{cor:err-prob-unif-1} implies the following uniform bound with high probability:
\bes{
\sup_{f \in \cF} \nm{f - \hat{f}}_{L^2_{\varrho}(D)} \lesssim E_{\cP}(\cF).
} 
Bounds of this type appear in \cite{temlyakov2021optimal}. See also \cite{krieg2026sampling} for extensions to other $L^p$-norm.
As we discuss in \S \ref{s:further}, this has implications in the study of \textit{sampling numbers} in information-based complexity and the efficacy of pointwise samples (so-called \textit{standard information}).
}

\rem{[The term $C_w$]
\label{rem:cw-term}
As an alternative to assuming that $w(x) \lesssim 1$, a.e.\ $x \sim \varrho$, one may also bound the term $C_w$ by assuming that $\cP$ contains a function $h$ with $\nm{h}_{L^2_{\varrho}(D)} = 1$ and $h(x) \gtrsim 1$, a.e.\ $x \sim \varrho$. This holds, for example, whenever the constant function $1 \in \cP$. In this case,
\bes{
C_w \lesssim \nms{h}_{\mathsf{disc},w} \leq \beta_w \nms{h}_{L^2_{\varrho}(D)} \leq \sqrt{1+\delta}.
}
However, as noted in Remark \ref{rem:noise-bound}, we can always construct $w$ so that the former assumption holds.
}

We now present a nonuniform bound that provides $L^2_{\varrho}$-quasi-optimality in probability, at the expense of a poor scaling with respect to the failure probability $\epsilon$. The proof is based on Markov's inequality, which, roughly speaking, is used to bound the discrete error term arising in \ef{LSerrbd}.

\cor{
[First nonuniform error bound in probability]
\label{cor:err-prob-1}
Let $0 < \delta,\epsilon < 1$, $f \in L^2_{\varrho}(D)$, $\cP \subset L^2_{\varrho}(D)$ be a finite-dimensional subspace with $\dim(\cP) = n$ and $\mu_1,\ldots,\mu_m$ be probability measures satisfying Assumption \ref{ass:mixture}. Consider sample points drawn randomly and independently with $x_i \sim \mu_i$, $i = 1,\ldots,m$, where 
\be{
\label{m-cond-one}
m \geq C_{\delta} \cdot \kappa_w(\cP) \cdot \log(4 n /\epsilon),\qquad C_{\delta} = ((1+\delta) \log(1+\delta) - \delta))^{-1},
}
and $w = 1/\nu$ and $\kappa_w$ are as in \ef{mu_weight_fn} and \ef{kappa-w-def}, respectively. Then the following hold with probability at least $1-\epsilon$. For any $e \in \bbC^m$, the weighted least-squares approximation $\hat{f}$ is unique and satisfies
\be{
\label{ls-err-bd-prob-1}
\nm{f - \hat{f}}_{L^2_{\varrho}(D)} \leq \left ( 1 + \sqrt{\frac{2 \kappa_w(\cP)}{m \epsilon}} \frac{1}{1-\delta} \right )\inf_{p \in \cP} \nm{f - p}_{L^{2}_{\varrho}(D)} + \frac{1}{\sqrt{1-\delta}} \nms{e}_{2,w}.
}
Moreover, if $\{ \phi_i \}^{n}_{i=1}$ is an orthonormal basis of $\cP$, then $\mathrm{cond}(A) \leq \sqrt{\frac{1+\delta}{1-\delta}}$.
}

Suppose again that $\delta = 1/2$ and $w(x) \lesssim 1$, $\forall x$. Then this bound implies that
\bes{
\nm{f - \hat{f}}_{L^2_{\varrho}(D)} \lesssim  ( 1 + 1/\sqrt{\epsilon \log(4 n / \epsilon) } ) \inf_{p \in \cP} \nm{f - p}_{L^{2}_{\varrho}(D)} + \nms{e}_2 / \sqrt{m}.
}
While stable and $L^2_{\varrho}$-quasi-optimal, the scaling with respect to $\epsilon$ is unappealing. To obtain an $\epsilon$-independent bound, this suggests we either need $n$ to be exponentially large in $1/\epsilon$, or impose an additional constraint on $m$ that $m \gtrsim \kappa_w(\cP) / \epsilon$. Neither is a desirable outcome.

One possible way to circumvent this issue involves using Bernstein's inequality instead of Markov's inequality. This exploits the fact that the discrete seminorm in \ef{LSerrbd} is a sum of independent random variables with bounded variance. It therefore concentrates exponentially fast (in $m$) around its mean $\bbE  [ \nm{f - p}^2_{\mathsf{disc},w}  ]= \nm{f-p}^2_{L^2_{\varrho}(D)}$ (recall \ef{exp-sum-scaling}). This leads to the following result, which is also nonuniform.

\cor{
[Second nonuniform error bound in probability]
\label{cor:err-prob-2}
Consider the setup of Corollary \ref{cor:err-prob-1} with \ef{m-cond-one} replaced by 
\be{
\label{m-conds-in-prob-2}
m \geq C_{\delta} \cdot \kappa_w(\cP) \cdot \log(4n /\epsilon)\quad \text{and}\quad m \geq 2 \cdot k \cdot \log(4/\epsilon)
}
for some $k > 0$. Then the following hold with probability at least $1-\epsilon$. For any $e \in \bbC^m$, the weighted least-squares approximation $\hat{f}$ is unique and satisfies
\be{
\label{ls-err-bd-prob-2}
\nm{f - \hat{f}}_{L^2_{\varrho}(D)} \leq \left ( 1 + \sqrt{\frac{2}{1-\delta}}  \right ) \inf_{p \in \cP} \left \{ \nm{f - p}_{L^{2}_{\varrho}(D)} + \frac{\nm{\sqrt{w}(f-p)}_{L^{\infty}_{\varrho}(D)}}{\sqrt{k}} \right \}  + \frac{1}{\sqrt{1-\delta}} \nms{e}_{2,w}.
}
Moreover, if $\{ \phi_i \}^{n}_{i=1}$ is an orthonormal basis of $\cP$, then $\mathrm{cond}(A) \leq \sqrt{\frac{1+\delta}{1-\delta}}$.
}

This result asserts a mixed type of quasi-optimality, involving the $L^2_{\varrho}$-norm and a (weighted) $L^{\infty}_{\varrho}$-norm divided by the factor $\sqrt{k}$. Notice that the factor $\sqrt{w}$ can be removed whenever $w(x) \lesssim 1$, a.e.\ $x \sim \varrho$, as will be the case later. Therefore, consider, as in Remark \ref{rem:unif-nonunif}, a setting where the $L^{2}_{\varrho}$-norm best approximation error decays algebraically fast in $n = \dim(\cP)$.
As we noted therein, the $L^{\infty}_{\varrho}$-norm best approximation error often decays $\ord{\sqrt{n}}$ slower than the former. Hence, one may choose $k = n$ in \ef{m-conds-in-prob-2} to show that $\hat{f}$ achieves the same algebraic convergence rate in $L^{2}_{\varrho}$-norm as the $L^{2}_{\varrho}$-norm best approximation in $\cP$. Note that this approach has also been used in the related context of function approximation via compressed sensing in \cite{rauhut2016interpolation} and \cite[\S 7.6]{adcock2022sparse}.

\subsection{Error bounds in expectation}\label{ss:err-bds-exp}

To obtain error bounds in expectation, we need to modify the least-squares estimator to avoid the `bad' regime where the discrete stability constants can be poorly behaved. In this section, we proceed as in \cite[\S 2.2]{dolbeault2021optimal}, which is based on \cite{cohen2013stability,cohen2017optimal}. 

Let $\{ \phi_i \}^{n}_{i=1}$ be an orthonormal basis of $\cP$ and we notice that \ef{alpha-beta-delta} holds whenever
\bes{
\nm{G - I}_{2} \leq \delta,
}
where $G$ is the discrete \textit{Gram matrix}
\bes{
G = \left ( \ip{\phi_j}{\phi_k}_{\mathsf{disc},w} \right )^{n}_{j,k=1} \in \bbC^{n \times n}.
}
We now have the following bound.

\lem{
\label{l:exp-bounds-chi}
Let $0 < \delta < 1$, $f \in L^2_{\varrho}(D)$, $\cP \subset L^2_{\varrho}(D)$ be a finite-dimensional subspace with $\mathrm{dim}(\cP) = n$ and $\mu_1,\ldots,\mu_m$ be probability measures satisfying Assumption \ref{ass:mixture}. Consider sample points drawn randomly and independently with $x_i \sim \mu_i$, $i = 1,\ldots,m$.
Then
\bes{
\bbE \left [ \nm{f - \hat{f}}^2_{L^2_{\varrho}(D)} \chi_{\nms{G-I}_2 \leq \delta} \right ] \leq \left ( 1 + \frac{2}{(1-\delta)^2} \frac{ \kappa_w(\cP)}{m} \right ) \inf_{p \in \cP} \nms{f - p}^2_{L^2_{\varrho}(D)} + \frac{2}{1-\delta} \bbE\left[\nms{e}^2_{2,w} \right ],
}
where $\chi_E$ denotes the indicator function of an event $E$ and $w = 1/\nu$ and $\kappa_w$ are as in \ef{mu_weight_fn} and \ef{kappa-w-def}, respectively.
}

This lemma can be used to construct several estimators that are stable and quasi-optimal in expectation. The first is the \textit{conditioned estimator} \cite{cohen2017optimal}, which is defined as
\bes{
\hat{f}^{\mathsf{ce}} = \hat{f} \chi_{\nms{G - I}_2 \leq \delta}.
}
Computing this estimator requires one to evaluate
\bes{
\nms{G-I}_2 = \max \{ | 1 - \sigma^2_{\max}(A) | , | 1 - \sigma^2_{\min}(A) | \}  = \max \{ | 1 - \alpha^2_w | , | 1- \beta^2_w | \}.
}
Having done so one simply sets $\hat{f}^{\mathsf{ce}} = \hat{f}$ if $\nms{G - I}_2 \leq \delta$ and $\hat{f}^{\mathsf{ce}} = 0$ otherwise. 

The conditioned estimator has the disadvantage that it requires an orthonormal basis for $\cP$ to be known -- a property that may not hold in practice (see \S \ref{s:further}). This can be avoided by using a \textit{truncated estimator}. This approach assumes an a priori bound for $f$ of the form
\bes{
\nms{f}_{L^2_{\varrho}(D)} \leq \sigma,
}
for some known $\sigma \geq 0$. Define the truncation operator $\cT_{\sigma} : L^2_{\varrho}(D) \rightarrow L^2_{\varrho}(D)$ by
$$
\cT_{\sigma}(g) 
:= \min\left\{1, \frac{\sigma}{\|g\|_{L_{\varrho}^2(\cU)}}\right\} g
=
\begin{cases}
g & \text{if }\|g\|_{L^2_{\varrho}(\cU)} \leq \sigma, \\
\sigma g / \|g\|_{L_{\varrho}^2(\cU)} & \text{otherwise},
\end{cases}
\quad \forall g \in L^2_{\varrho}(\cU).
$$
Then the truncated estimator is given by
\bes{
\hat{f}^{\mathsf{te}} = \cT_\sigma(\hat{f}).
}
Note that one can also construct a truncated estimator with respect to other norms. For example, the $L^{\infty}_{\varrho}$-norm was used in \cite{cohen2013stability,dolbeault2021optimal}.

\thm{
[Nonuniform error bound in expectation]
\label{t:err-bd-exp}
Let $0 < \delta,\epsilon < 1$, $f \in L^2_{\varrho}(D)$, $\cP \subset L^2_{\varrho}(D)$ be a finite-dimensional subspace with $\mathrm{dim}(\cP) = n$ and $\mu_1,\ldots,\mu_m$ be probability measures satisfying Assumption \ref{ass:mixture}. Consider sample points drawn randomly and independently with $x_i \sim \mu_i$, $i = 1,\ldots,m$, where
\be{
\label{sampl-com-exp-err}
m \geq C_{\delta} \cdot \kappa_w(\cP) \cdot \log(2n/\epsilon), \qquad C_{\delta} = ((1+\delta) \log(1+\delta) - \delta))^{-1},
}
and $w = 1/\nu$ and $\kappa_w$ are as in \ef{mu_weight_fn} and \ef{kappa-w-def}, respectively.
Then
\bes{
\bbE \left[ \nm{f - \hat{f}^{\mathsf{ce}}}^2_{L^2_{\varrho}(D)} \right ]  \leq \left ( 1 + \frac{2}{(1-\delta)^2} \frac{ \kappa_w(\cP)}{m} \right ) \inf_{p \in \cP} \nms{f - p}^2_{L^2_{\varrho}(D)} + \frac{2}{1-\delta} \bbE \left [ \nms{e}^2_{2,w} \right ]+ \nms{f}^2_{L^2_{\varrho}(D)} \epsilon.
}
The same bound holds for $\hat{f}^{\mathsf{te}}$, except with the final term replaced by $4 \sigma^2 \epsilon$.
}

Observe that the factor
\bes{
\frac{2}{(1-\delta)^2} \frac{\kappa_w(\cP)}{m} \leq  \frac{2}{(1-\delta)^2 C_{\delta}} \frac{1}{\log(2n/\epsilon)} \rightarrow 0
}
as $n \rightarrow \infty$. Hence, this bound asserts $L^2_{\varrho}$-quasi-optimality of the two estimators (with constant approaching $1$ as $n \rightarrow \infty$) up to the $\ord{\epsilon}$ term in the error bound.

\rem{
[Removing the $\epsilon$ term]
\label{rem:removing-eps}
As discussed in \cite{dolbeault2021optimal}, this $\ord{\epsilon}$ term may be problematic. For example, suppose that the best approximation error decays geometrically fast in $n = \dim(\cP)$, i.e., $\inf_{p \in \cP } \nm{f-p}_{L^2_{\varrho}(D)} \asymp \rho^{-n}$  for some $\rho > 1$. Then achieving the same rate for the least-squares approximation in expectation would require setting $\epsilon = \ord{\rho^{-n}}$. This adds an additional multiplicative factor of $n$ to the sample complexity bound \ef{sampl-com-exp-err}, thus prohibiting a geometric rate of convergence in terms of $m$ (recall that $\kappa_w(\cP) \geq n$). One way to remove this term, which was introduced in \cite{haberstich2022boosted}, is to repeatedly redraw the sample points $\{x_1,\ldots,x_m\}$ until the condition $\nm{G - I}_2 \leq \delta$ is met, and then use the resulting points to construct a weighted least-squares estimator $\hat{f}^{\star}$. This is another type of conditioned estimator, but it is not the same as the estimator $\hat{f}^{\mathsf{ce}}$ considered above. However, by using $\hat{f}^{\star}$ one can achieve a similar error bound in expectation, except without the $\ord{\epsilon}$ term (the parameter $\epsilon$ now only influences the expected number of redraws needed to achieve $\nm{G - I}_2 \leq \delta$). See \cite{haberstich2022boosted,dolbeault2021optimal,dolbeault2022optimal} for further information.
}

\rem{
[The noise bound]
\label{rem:noise-bound-2}
Because of Assumption \ref{ass:mixture}, the noise term in Theorem \ref{t:err-bd-exp} satisfies the bound
\bes{
\bbE \left[ \nm{e}^2_{2,w} \right ]= \frac1m \sum^{m}_{i=1} \bbE \left [w(x_i) | e_i|^2 \right ] = \frac1m \sum^{m}_{i=1} \int_{D} w(x) \D \mu_i(x) | e_i|^2 \leq  \nm{e}^2_{\infty}.
}
If $\mu_1 = \cdots = \mu_m = \mu$, then one has the precise expression $\bbE [ \nm{e}^2_{2,w}  ]= \frac1m \nm{e}^2_2$. In general, one also the bound $\bbE [ \nm{e}^2_{2,w}  ] \lesssim \frac1m \nm{e}^2_2$ whenever $w(x) \lesssim 1$, a.e. $x \sim \varrho$, as in Remark \ref{rem:noise-bound}.
}

\section{Christoffel sampling}\label{s:near-opt}

We now come to the crux of this article, which is to devise random sampling schemes that achieve near-optimal sample complexity bounds.

\subsection{Optimal choice of weight function via the Christoffel function}\label{ss:near-opt-samp}

The results shown in the previous section relate the number of measurements $m$ to the constant $\kappa_w(\cP)$. Hence, our goal is to choose a weight function $w$ that minimizes this constant. Recall that  
\bes{
\kappa_w(\cP) = \esssup_{x \sim \varrho} w(x) \cK(\cP)(x).
}
A natural first choice involves selecting 
\bes{
w(x) \propto \frac{1}{\cK(\cP)(x)}.
}
Applying the normalization condition \ef{w_normalization} (recall that $\nu = 1/w$) and the fact that $\int_{D} \cK(\cP)(x) \D \varrho(x) = n$ (recall \ef{Kappa-def-alt}), we obtain
\be{
\label{w-opt-orig}
w^{\star}(x) = \frac{n}{\cK(\cP)(x)}.
}
This choice is quite popular in the literature. 
However, it requires the additional assumption that $\cK(\cP)(x) > 0$ almost everywhere. This is a rather mild assumption, which is equivalent to requiring that for almost every $x \sim \varrho$ there exists a $p = p_x \in \cP$ for which $p_x(x) \neq 0$ (in particular, it is implied by the assumption made in Remark \ref{rem:cw-term}). If this holds, then \ef{w-opt-orig} is the optimal for choice of $w$, since $\kappa_{w^{\star}}(\cP) = n$ achieves the optimal lower bound \ef{kappa-w-lb} for $\kappa_{w}(\cP)$. 

However, this choice may not be desirable, for the reasons considered in  Remark \ref{rem:noise-bound}. Fortunately, this issue can be resolved, and the above assumption avoided, by modifying $w^{\star}(x)$ to
\be{
\label{w-opt}
w^{\star}(x) = \left ( \frac{1}{2} + \frac12 \frac{\cK(\cP)(x)}{n} \right )^{-1}.
}
This leads to a positive and bounded weight function satisfying $w^{\star}(x) \leq 2$.
The only cost is suboptimality by a factor of at most $2$, i.e., $\kappa_{w^{\star}}(\cP) \leq 2n$.
The reader will likely notice that one could replace the $1/2$ in \ef{w-opt} with a weighted combination $\theta + (1-\theta) \frac{\cK(\cP)(x)}{n}$ for any $0 < \theta < 1$, giving $\kappa_{w^{\star}}(\cP) \leq (1-\theta)^{-1} n$ and $w^{\star}(x) \leq \theta^{-1}$. For simplicity, we consider the factor $1/2$ throughout.  The construction \ef{w-opt} was first considered by \cite{pozharska2022note}, although, as commented therein, it was also used implicitly in several earlier works.

This aside, having chosen $w = w^{\star}$ as in \ef{w-opt-orig} (one could also consider \ef{w-opt}), to achieve near-optimal sampling we need to select sampling measures $\mu_1,\ldots,\mu_m$ such that \ef{mu_weight_fn} holds, i.e.,
\be{
\label{mu_weight_fn_opts}
\frac1m \sum^{m}_{i=1} \D \mu_i(x) = \frac{\cK(\cP)(x)}{n}  \D \varrho(x)=\frac{\sum^{n}_{i=1} | \phi_i(x) |^2 }{n}  \D \varrho(x).
}
In this case, the various sample complexity estimates of the previous section are \textit{near-optimal} in $n$. Indeed, letting $\delta = 1/2$, we see that the condition
\bes{
m \gtrsim n \cdot \log(2n/\epsilon)
}
guarantees the various `in probability' or `in expectation' bounds presented in the previous section.

\subsection{Christoffel sampling}

There are several ways to achieve \ef{mu_weight_fn_opts}. Arguably the most popular is
\be{
\label{mu-opt-1}
\mu^{\star}_1 = \cdots = \mu^{\star}_m = \mu ,\qquad \text{where }\D \mu^{\star}(x) =   \frac{\cK(\cP)(x)}{n} \D \varrho(x) = \frac{\sum^{n}_{i=1} | \phi_i(x) |^2 }{n} \D \varrho(x).
}
However, this strategy is not well suited in the case of hierarchical approximation (Remark \ref{rem:hierarchical}). Indeed, for $k \in \bbN$, let $\mu^{(k),\star}$ be given by \ef{mu-opt-1} for $\cP = \cP^{(k)}$. Suppose that the first $m_1$ points $x^{(1)}_i \sim_{\mathrm{i.i.d.}} \mu^{(1),\star}$, $i = 1,\ldots,m_1$. Then we would like to re-use these $m_1$ points $\{ x^{(1)}_i \}^{m_1}_{i=1}$ when constructing the second set of sample points $\{ x^{(2)}_i \}^{m_2}_{i=1}$. However, since $\mu^{(1),\star} \neq \mu^{(2),\star}$ in general, these $m_1$ points are drawn with respect to the wrong measure for near-optimal sampling in the subspace $\cP^{(2)}$. Thus, it is not clear how to achieve near-optimal sampling simply by augmenting the set $\{ x^{(1)}_i \}^{m_1}_{i=1}$ with $m_2 - m_1$ new points.
 
One strategy to overcome this limitation involves interpreting $\mu^{(k+1),\star}$ as an additive mixture of $\mu^{(k),\star}$ and a certain update measure $\sigma^{(k),\star}$. One can then use this to construct a sampling procedure that recycles `most' of the first $m_1$ points, while ensuring that the overall sample is drawn i.i.d.\ from $\mu^{(k+1)}$ \cite{arras2019sequential}.

An alternative approach, introduced in \cite{migliorati2019adaptive}, involves choosing measures $\mu_i$ according to the individual basis functions. For simplicity, consider a single subspace $\cP$. Let $\{ \phi_i \}^{n}_{i=1}$ be an orthonormal basis of $\cP$ and suppose that $m = r n$ for some $r \in \bbN$. Then we define
\be{
\label{mu-opt-2}
\D \mu^{\star}_i(x) =  | \phi_j(x) |^2 \D \varrho(x),\quad (j-1) r < i \leq j r,\ j = 1,\ldots,n.
}
Observe that
\bes{
\frac1m \sum^{m}_{i=1} \D \mu^{\star}_i(x) =  \frac{r}{m} \sum^n_{j=1} | \phi_j(x) |^2 \D \varrho(x) = \frac{\cK(\cP)(x)}{n} \D \varrho(x),
}
due to \ef{Kappa-def-alt}. Hence \ef{mu_weight_fn_opts} also holds for this choice, guaranteeing near-optimal sample complexity. Crucially, each sampling measure corresponds to a single basis function, rather than an additive mixture of basis functions as in \ef{mu-opt-1}. Therefore this approach readily lends itself to hierarchical approximation. See \cite{adcock2020near-optimal,migliorati2019adaptive} for further details. Note that the distributions \ef{mu-opt-2} are known as \textit{induced distributions} \cite{narayan2018computation,guo2020constructing}, as they are induced by the orthonormal basis $\{ \phi_i \}^{n}_{i=1}$. 

Henceforth, we will refer to either procedure -- or, indeed, any selection of measures $\mu^{\star}_i$ for which \ef{mu_weight_fn} holds for $w^{\star} = 1/\nu^{\star}$ as in \ef{w-opt-orig} or \ef{w-opt} -- as \textit{Christoffel sampling}.

\subsection{A further uniform error bound in probability}\label{ss:unif-prob}

To conclude this section, we now describe how a further modification of the near-optimal sampling measure can lead to uniform bounds in probability that improve on the somewhat crude bounds shown in Corollary \ref{cor:err-prob-unif-1} and achieve something close to $L^2_{\varrho}$-quasi-optimality. This section is based on techniques developed in \cite{krieg2021function,krieg2021functionII} to estimate \textit{sampling numbers}. See \S \ref{s:further} for additional discussion.

We now assume that there is an orthonormal basis $\{ \phi_i \}^{\infty}_{i=1} \subset L^2_{\varrho}(D)$ and that
\be{
\label{Pn-def-unif}
\cP = \cP_n = \spn \{ \phi_1,\ldots,\phi_n\}.
}
For convenience, given $f \in L^2_{\varrho}(D)$ let
\be{
\label{e-def}
e_n(f) = \inf_{p \in \cP_n} \nms{f - p}_{L^2_{\varrho}} = \sqrt{\sum^{\infty}_{i > n} | c_i |^2},
}
where $c_i = \ip{f}{\phi_i}_{L^2_{\varrho}(D)}$ is the $i$th coefficient of $f$. The second equality is due to Parseval's identity.
We now construct the sampling measure. Define sets
\bes{
I_l = \{ n 2^{l} + 1,\ldots, n 2^{l+1} \},\quad l = 0,1,2,\ldots 
}
and consider a sequence $(v_l)^{\infty}_{l=0}$ with $\sum^{\infty}_{l=0} v^2_l = 1$. Then we set
\be{
\label{opt-meas-krieg}
\D \mu^{\star}(x) = \left ( \frac12 + \frac14 \frac{\sum^{n}_{i=1} | \phi_i (x) |^2}{n} + \frac14 \sum^{\infty}_{l=0} \frac{v^2_l}{|I_l|} \sum_{i \in I_l} | \phi_i(x) |^2 \right ) \D \varrho(x).
}

\thm{
[Second uniform error bound in probability]
\label{thm:ullrich-approach}
Let $0 < \delta,\epsilon < 1$, $n \in \bbN$, $\{ \phi_i \}^{\infty}_{i=1} \subset L^2_{\varrho}(D)$ be an orthonormal basis and $\cP = \cP_n$ be as in \ef{Pn-def-unif}. Let $0 < p < 2$ and $v_l = C_{\theta} 2^{-\theta l}$ for $0 < \theta < 1/p-1/2$, where $C_\theta$ is such that $\sum^{\infty}_{l=0} v^2_l = 1$. Consider sample points $x_i \sim_{\mathrm{i.i.d.}} \mu^{\star}$, $i=1,\ldots,m$, where $\mu^{\star}$ is as in \ef{opt-meas-krieg} and
\be{
\label{m-est-krieg}
m \geq 4 \cdot C_{\delta} \cdot n \cdot \log(4n/\epsilon),\qquad \text{where }C_{\delta} = ((1+\delta) \log(1+\delta)-\delta))^{-1}.
}
Then the following holds with probability at least $1-\epsilon$. For any $f$ that is defined everywhere in $D$ and for which $(e_n(f) )^{\infty}_{n=1} \in \ell^p(\bbN)$, and any noise $e \in \bbC^m$, the weighted least-squares approximation $\hat{f}$ is unique and satisfies
\be{
\label{f-hatf-bound}
 \nm{f - \hat{f}}_{L^2_{\varrho}(D)} \leq \frac{C_{p,\theta}}{\sqrt{1-\delta}} \left [ e_n(f) + \left ( \frac{1}{n} \sum_{k > n} (e_k(f))^p \right )^{1/p}  \right ] + \sqrt{\frac{2}{1-\delta}} \frac{\nms{e}_2}{\sqrt{m}},
}
where $C_{p,\theta} > 0$ is a constant depending on $p$ and $\theta$ only. Moreover, the condition number of the least-squares matrix \ef{ls-Ab} satisfies $\mathrm{cond}(A) \leq \sqrt{\frac{1+\delta}{1-\delta}}$.
}

To put this result into context, consider a class $\cK$  of functions that are defined everywhere on $D$ and for which 
\bes{
\sup_{f \in \cK} e_n(f) \asymp n^{-\alpha} \log^{\beta}(n+1)
} 
for some $\alpha > 1/2$ and $\beta \in \bbR$. This holds, for instance, in the case of polynomial approximation when $\cK$ is a unit ball of functions of finite regularity. Then the error term on the right-hand side of \ef{f-hatf-bound} satisfies
\bes{
e_n(f) + \left ( \frac{1}{n} \sum_{k > n} (e_k(f))^p \right )^{1/p} \leq C_{\alpha,\beta,p} n^{-\alpha} \log^{\beta}(n+1),\quad \forall p \in (1/\alpha,2).
}
Hence, with probability at least $1-\epsilon$, one obtains a matching error decay rate for the least-squares estimator (up to constants), uniformly for functions $f \in \cK$, with near-optimal sample complexity.

However, it is unclear how to compute this approximation in practice. As we discuss in \S \ref{s:further}, implementing standard Christoffel sampling may itself not always be straightforward. The sampling measure \ef{opt-meas-krieg} also has the additional complication that it involves an infinite sum. 

\subsection{Summary}

We summarize this section as follows. Christoffel sampling involves choosing sampling measures according to the Christoffel function of the space $\cP$. It leads provably to near-optimal, log-linear sample complexity. These measures can always be designed to ensure a stable approximation, and can chosen in such a way to facilitate hierarchical approximation schemes. Further modifications also allow for stronger uniform error bounds in probability.

\section{Improvement over Monte Carlo sampling}\label{ss:MC-sampling}

Having introduced Christoffel sampling, in this section we discuss how it compares against standard Monte Carlo sampling, i.e., i.i.d.\ random sampling from the measure $\varrho$. Recall that in this case the weight function $w$ is precisely $w \equiv 1$, meaning that Monte Carlo sampling corresponds to an unweighted least-squares approximation. Theorem \ref{t:alpha-beta-est} asserts that its sample complexity depends on the unweighted quantity
\be{
\label{MC-kappa}
\kappa(\cP) = \kappa_1(\cP) = \nms{\cK(\cP)}_{L^{\infty}_{\varrho}(D)}.
}
In other words, the maximal behaviour of the Christoffel function $\cK(\cP)$, or equivalently, the optimal constant in the unweighted Nikolskii-type inequality \ef{nikolskii}.

\subsection{Bounded orthonormal systems}

Recall that $\kappa_w(\cP) \geq n$ for any $w$, and therefore, $\kappa(\cP) \geq n$. If $\kappa(\cP) \leq C n$ for some $C \geq 1$, then Monte Carlo sampling is already a near-optimal strategy, and there may be little need to optimize the sampling measure (besides reducing the constant $C$).
This situation occurs whenever $\cP$ has an orthonormal basis $\{ \phi_i \}^{n}_{i=1}$ that is \textit{uniformly bounded}. Such a basis is sometimes referred to as a \textit{bounded orthonormal system} (see, e.g., \cite[Chpt.\ 12]{foucart2013mathematical} or \cite[Chpt.\ 6]{adcock2022sparse}). Specifically, if
\be{
\label{BOS-c}
\nms{\phi_i}^2_{L^{\infty}_{\varrho}(D)} \leq C,\quad \forall i = 1,\ldots,n,
}
then it follows immediately from \ef{Kappa-def-alt} that $\kappa(\cP) \leq C n$. Subspaces of trigonometric polynomials are a standard example for which this property holds (with $C = 1$). Closely related are the Chebyshev polynomials of the first kind (see \S \ref{ss:orth-polys}). The orthonormal Chebyshev polynomials on $(-1,1)$ are defined by
\bes{
\psi_{0}(x) = 1,\quad \psi_{i}(x) = \sqrt{2} \cos(i \arccos(x)),\ i \in \bbN.
}
Therefore they satisfy \ef{BOS-c} with $C = 2$. In $d$ dimensions, the tensor-product Chebyshev polynomial basis on $(-1,1)^d$ satisfies \ef{BOS-c} with $C = 2^d$. Hence it is also a bounded orthonormal system, albeit with a constant that grows exponentially fast as $d \rightarrow \infty$. 

\subsection{Arbitrarily-bad sample complexity bounds}

Unfortunately, bounded orthonormal bases are quite rare in practice. It is also not uncommon to encounter subspaces $\cP$ for which $\kappa(\cP)$ can be arbitrarily large in comparison to $n$. This occurs, for instance, when considering Legendre polynomials. Unlike the Chebyshev polynomials, the univariate Legendre polynomials grow with their degree, and satisfy (see, e.g., \cite[\S 2.2.2]{adcock2022sparse})
\be{
\label{leg-poly-unif-bd}
\nm{\psi_i }_{L^{\infty}(-1,1)} = | \psi_i(\pm1) | = \sqrt{i+1/2}.
}
Therefore, for any set $S \subset \bbN_0$, the space $\cP_{S} = \spn \{ \psi_i : i \in S \}$ has constant $\kappa(\cP_S)$ given by $\kappa(\cP_S) = \sum_{i \in S} (i+1/2)$ (this follows from \ef{Kappa-def-alt} and \ef{MC-kappa}). By choosing the entries in $S$ arbitrarily large, the following result is now immediate.

\prop{
\label{prop:arbitrarily-bad}
There exists a probability space $(D,\cD,\varrho)$ such that the following holds. For every $n \in \bbN$ and $C > 0$, there exists a subspace $\cP \subset L^2_{\varrho}(D)$ of dimension $n$ such that $\kappa(\cP) \geq C$.
}

The reason for this poor behaviour is that Legendre polynomials are highly peaked near $x = \pm 1$. The functions $\psi_i$ are $\ord{1}$ as $i \rightarrow \infty$ uniformly within compact subsets of $(-1,1)$, yet, as noted in \ef{leg-poly-unif-bd}, they behave like $\ord{\sqrt{i}}$ at the endpoints. This points towards a general observation. We expect poor scaling of $\kappa(\cP)$, and therefore poor performance of Monte Carlo sampling, whenever $\cP$ contains a function that is highly localized.

\subsection{Sample complexity bounds for polynomial spaces}

Following \S \ref{s:multivar-poly}, we now present a series of bounds for $\kappa(\cP)$ in the case of multivariate polynomial spaces $\cP = \bbP_{S}$, focusing on the case where $S$ is a lower set.

\pbk
\textit{Chebyshev polynomials.} As noted previously, any subspace
\bes{
\cP_S = \spn \{ \Psi_{\nu} : \nu \in S \},\qquad S \subset \bbN^d_0,\ |S| = n,
}
of multivariate Chebyshev polynomials satisfies the exponentially-large (in $d$) bound $\kappa(\cP_S) \leq 2^d n$. Fortunately, if $S$ is a lower set, then one has the $d$-independent bound (see, e.g., \cite[Prop.\ 5.13]{adcock2022sparse})
\be{
\label{cheb-lower-S}
\kappa(\cP_S) = \kappa(\bbP_S) \leq n^{\log_2(3)}.
}
This estimate is sharp up to a constant, i.e., there is a lower set with $\kappa(\cP_S) \gtrsim n^{\log_2(3)}$. It is also tighter than the previous bound  $\kappa(\cP_S) \leq 2^d n$ whenever $d \geq \log_2(3/2) \log_2(n)$. 

Improvements such as this are typical when lower set structure is imposed. We will see another instance of this in a moment. It is one of the features that makes lower sets desirable for high-dimensional approximation -- the underlying reason being that a lower set cannot contain too many `large' (or even nonzero) indices. 

\pbk
\textit{Legendre polynomials.} Because of \ef{Kappa-def-alt}, \ef{MC-kappa} and \ef{leg-poly-unif-bd}, for any subspace $\cP_S = \spn \{ \Psi_{\nu} : \nu \in S \}$ of Legendre polynomials one has
\bes{
\kappa(\cP_S) = \sum_{\nu \in S} \prod^{d}_{k=1} (\nu_k+1),
}
which can be arbitrarily large in comparison to $n = |S|$. However, imposing lower set structure leads to a dramatic improvement. If $S$ is lower then the following sharp upper bound holds (see, e.g., \cite[Prop.\ 5.17]{adcock2022sparse}):
\be{
\label{kappa-legendre}
\kappa(\cP_S) \leq n^2.
}

\pbk
\textit{Ultraspherical and Jacobi polynomials.} The situation for Jacobi polynomials with $\max \{ \alpha,\beta \} > -1/2$ is similar to that of Legendre polynomials. The quantity $\kappa(\cP_S)$ can be arbitrarily large for general $S$. However, if $S$ is lower then one has the finite bound
\be{
\label{kappa-jacobi}
\kappa(\cP_S) \leq n^{2\max \{ \alpha , \beta \} + 2 },\qquad \forall \alpha,\beta \in \bbN_0,
}
for Jacobi polynomials (see \cite[Thm.\ 9]{migliorati2015multivariate}) and
\be{
\label{kappa-ultraspherical}
\kappa(\cP_S) \leq n^{2 \alpha + 2 },\qquad \forall 2\alpha+1 \in \bbN,
}
for ultraspherical polynomials  (see \cite[Thm.\ 8]{migliorati2015multivariate}).

\rem{
[Sharpness of the rates]
The bounds \ef{kappa-legendre}--\ef{kappa-ultraspherical} imply that a superlinear sample complexity suffices for stable and accurate polynomial approximation with random samples drawn from Jacobi measures. These rates are also necessary. This was recently shown in \cite{xu2024stability} in the $d = 1$ case, based on earlier work \cite{platte2011impossibility,adcock2019optimal} on deterministic points that are equidistributed with respect to such measures. Specifically, \cite{xu2024stability} shows that choosing $m \asymp n^{\tau}$, where $\tau < 2(\max \{ \alpha,\beta\}+1)$, necessarily implies exponential instability of the least-squares approximation (or, indeed, any other approximation method that achieves similar accuracy in the noiseless setting).
}

\noindent
\textit{Bounded, non-tensor product domains.} Several of these bounds generalize to bounded non-tensor product domains \cite{prymak2019christoffel,ditzian2016nikolskii,adcock2020approximating,dolbeault2021optimal}. If $\varrho$ is the uniform measure and $D \subset \bbR^d$ is bounded and satisfies the \textit{inner cone} condition (see, e.g., \cite[\S 4.11]{adams2003sobolev}), then
\bes{
\kappa(\cP_S) \leq C_D \cdot n^2,\quad \text{if $S = S^{\mathsf{TD}}_p$ and $n = | S^{\mathsf{TD}}_p|$.}
}
See \cite[Thm.\ 5.4]{dolbeault2021optimal}.
However, as shown in  \cite[Thm.\ 5.6]{dolbeault2021optimal}, when $D$ has $C^2$ boundary one also has the sharper bound
\bes{
\kappa(\cP_S) \leq C_D \cdot n^{\frac{d+1}{d}},\quad \text{if $S = S^{\mathsf{TD}}_p$ and $n = | S^{\mathsf{TD}}_p|$.}
}
See \cite{dolbeault2021optimal} and references therein for further results of this type. These bounds apply only to the total degree index set \ef{TP-TD}. For arbitrary lower sets, one has
\bes{
\kappa(\cP_S) \leq n^2 /\lambda,\quad \text{if $S$ is lower and $n = |S|$,}
}
whenever $D$ has the \textit{$\lambda$-rectangle property}: namely, there is a $0 < \lambda < 1$ such that, for any $x \in D$, there exists an axis-aligned rectangle $R \subseteq D$ containing $x$ for which $|R| \geq \lambda |D|$. See \cite[Thm.\ 2.4]{adcock2020approximating}.

\pbk
\textit{Hermite and Laguerre polynomials.} Unfortunately, this analysis of Monte Carlo sampling says nothing about Hermite and Laguerre polynomial approximations, for the simple reason that such polynomials are not uniformly bounded, and therefore the constant \ef{MC-kappa} satisfies $\kappa(\cP) = +\infty$. The sample complexity of Hermite and Laguerre polynomial approximation is poorly understood in comparison to that of Jacobi polynomials. A number of empirical studies have suggested a \textit{super-polynomial} or \textit{exponential} sample complexity (in $n$, for fixed $d$). But relatively few theoretical estimates exist. See \cite{migliorati2013polynomial,guo2020constructing,hampton2015coherence,tang2014discrete} and \cite[Rem.\ 5.18]{adcock2022sparse}. Suffice to say, though, Hermite and Laguerre polynomial approximations are examples where one stands to gain substantially from Christoffel sampling, and as such, these have often been used as examples to illustrate its efficacy \cite{guo2020constructing,hampton2015coherence,cohen2017optimal}.

\subsection{Summary}

In summary, in this section we have shown, firstly, that the sample complexity of Monte Carlo sampling depends on the $L^{\infty}_{\varrho}$-norm of the Christoffel function, and, secondly, that it is easy to construct case where this quantity is arbitrarily-large in comparison to $n = \dim(\cP)$. Moreover, these scenarios arise in various polynomial approximation problems, especially when considering unbounded domains. Thus, Monte Carlo sampling can be arbitrarily bad in comparison to Christoffel sampling.

\section{Variations, extensions and further topics}\label{s:further}

To conclude this first part of the article, we now discuss a number of issues not considered so far, along with some variations and extensions.

\subsubsection*{Sampling from the near-optimal measures}\label{ss:practical-sampling}

A key practical issue is drawing samples from the measure(s) \ef{mu-opt-1} or \ef{mu-opt-2} in a computationally-efficient manner. If the Christoffel function $\cK(x)$ can be evaluated at any point $x$, then a first option, as studied in \cite{hampton2015coherence}, is to use Markov Chaine Monte Carlo (MCMC) techniques such as Metropolis--Hastings. However, reliable computations with MCMC can be challenging. For instance, they rely on a good choice of proposal distribution and careful tuning of various parameters.

If an orthonormal basis $\{ \phi_i \}^{n}_{i=1}$ for $\cP$ is known, then one can always evaluate $\cK(x)$ via \ef{Kappa-def-alt} and use MCMC. However, many problems also have a tensor-product structure -- for example, the problem considered in \S \ref{ss:orth-polys}, which involves tensor products of orthogonal polynomials -- and this can be used to design improved algorithms. Two options considered in \cite{cohen2017optimal} are \textit{sequential conditional sampling} and \textit{mixture sampling}. The former relies on the fact that the multivariate density \ef{mu-opt-1} can be written as a product of univariate conditional densities, which are known explicitly in the tensor-product case. The latter exploits the fact that  \ef{mu-opt-1} is an additive mixture of induced distributions \ef{mu-opt-2}, which involve tensor-product densities. Thus, both approaches reduce the problem to sampling from certain univariate densities, which can be done, for instance, by rejection sampling or inverse transform sampling. In particular, efficient algorithms for sampling from the induced distributions of classical orthogonal polynomials have been introduced in \cite{narayan2018computation}.

Unfortunately, many problems either do not have a tensor-product structure or may lack explicit orthonormal bases. This arises, for example, in multivariate polynomial approximation on irregular domains. Orthogonal polynomials can be defined explicitly for certain non-tensorial domains, e.g., simplices, balls and spheres \cite{dunkl2014orthogonal}. Yet, this is impossible in general. In cases such as these, where even evaluating the Christoffel function may not be straightforward, a simple, albeit crude approach is to use a grid \cite{adcock2020near-optimal,migliorati2021multivariate}. Let
$D$ be a grid $Z = \{ z_i \}^{K}_{i=1}$ of points drawn i.i.d.\ from $\varrho$ and consider the discrete uniform measure $\bar{\varrho} = K^{-1} \sum^{K}_{i=1} \delta_{z_i}$ supported on $Z$. Given a nonorthogonal spanning set $\{ \psi_i \}^{n}_{i=1}$ for $\cP$, now considered a subspace of $L^2_{\bar{\varrho}}(D)$, one may construct an orthonormal basis via numerical linear algebra tools such as classical QR factorizations or SVDs, or through more recent approaches such as \textit{V+A (Vandermonde with Arnoldi)} \cite{zhu2023convergence}. Sampling from the measure(s) \ef{mu-opt-1} or \ef{mu-opt-2} is now straightforward, since it is now a discrete measure supported on $Z$.

This approach, which is a form of leverage score sampling (see next), ensures accurate and stable recovery in the $L^2_{\bar{\varrho}}$-norm. To guarantee the same in the original $L^2_{\varrho}$-norm one needs to choose $K$ large enough. Since $Z$ is obtained by Monte Carlo sampling, this is governed by the constant \ef{MC-kappa}. This is another reason why the results of \S \ref{ss:MC-sampling} are important, since they inform the size of grid needed in such a construction. See \cite{adcock2020near-optimal,migliorati2021multivariate} for further details.  

Note that the computation of an orthonormal basis over $Z$ is a purely offline cost. It does not involve additional evaluations of the target function $f$, which are often the main computational bottleneck in practice.  Of course, theoretical bounds for the Christoffel function may not be available or, if available, may result in a value of $K$ that is too large for practical computations. This has spurred several recent works which introduced more refined strategies for constructing the grid $K$ than vanilla Monte Carlo sampling \cite{trunschke2024optimal2,dolbeault2021optimal}. Another alternative is to use a structured grid for $Z$. See \cite{bartel2024reconstruction} for an approach based on Quasi-Monte Carlo grids for least-squares approximation with the Fourier basis. This has the additional benefit that the algebraic least-squares problem can be solved efficiently via Fast Fourier Transforms (FFTs).

\subsubsection*{Connections to matrix sketching and leverage score sampling}

Let $X \in \bbC^{N \times n}$, where $N \geq n$, and $x \in \bbC^N$. In applications in data science, it may be infeasible to solve the `full' least squares problem $w \in \argmin{z \in \bbC^n} \nm{X z - x}_{2}$. \textit{Matrix sketching} involves using a sketching matrix $S \in \bbC^{m \times N}$ (a matrix with one nonzero per row) and solving the sketched problem $\hat{w} \in \argmin{z \in \bbC^n} \nm{S X z - S x}_2$. The objective is to find $S$ with a minimal number of rows $m$ such that
\bes{
\nm{X \hat{w} - x}_{\ell^2} \lesssim \nm{X w - x}_{2}.
}
In \textit{random} sketching, one considers a discrete probability distribution $\pi = \{ \pi_1,\ldots,\pi_N \}$ on $\{1,\ldots,N\}$ with $\pi_i > 0$, $\forall i$, draws $j_1,\ldots,j_m \sim_{\mathrm{i.i.d.}} \pi$ and then sets $S_{i,j_i} = 1/\sqrt{\pi_{j_i}}$ and $S_{ij} = 0$ otherwise. A near-optimal choice of the distribution $\pi$ is provided by the \textit{leverage scores} $\ell_i(X)$, $i = 1,\ldots,N$, of the matrix $X$. These are given by
\be{
\label{leverage-scores}
\ell_i(X) = \max_{\substack{z \in \bbC^n \\ X z \neq 0}} \frac{|(X z)_i|^2}{\nm{X z}^2_2} \equiv \nm{Q(i,:)}^2_2, 
}
where $Q \in \bbR^{N \times n}$ is any matrix whose columns form an orthonormal basis for $\mathrm{Ran}(X)$.
The resulting procedure is the well-known technique of \textit{leverage score sampling} \cite{drineas2012fast,mahoney2011randomized,woodruff2014sketching}.

Leverage score sampling can be considered as a special case of Christoffel sampling involving the discrete set $D = \{1,\ldots,N\}$. Note that any vector $x = (x_i)^{N}_{i=1} \in \bbC^N$ can be viewed as a function $f \in L^2_{\varrho}(D)$ via the relation $f(i) = x_i$ and vice versa. One now defines the subspace $\cP = \{ X z : z \in \bbC^n \}$ to cast the above problem into the form introduced in \S \ref{s:prelims}. In particular, the values of the Christoffel function $\cK(\cP)$ over the set $D$ are precisely the leverage scores \ef{leverage-scores} of the matrix $X$. We refer to \cite[Sec.\ A.2]{adcock2023cs4ml} and \cite{malik2022fast} for details.

\subsubsection*{$\ord{n}$ sampling strategies, frame subsampling and Kadison--Singer}

The Christoffel sampling schemes described in this paper are near-optimal, in the sense that the sample complexity is log-linear in $n$. Due to the coupon collector's problem, this is the best achievable when using i.i.d.\ samples (see, e.g., \cite[Rem.\ 3]{dolbeault2022optimal}).

This limitation has spurred a recent line of work on methods that have linear sample complexity in $n$, i.e., optimal up to possible constants, and even schemes that can achieve interpolation, i.e., $m = n$. This work is based on techniques introduced in \cite{marcus2015interlacing,batson2014twice} on \textit{frame subsampling} (later, we also discuss a different approach in the hierarchical setting based on randomized quadratures). Here, given a frame of $m \geq n$ vectors in $\bbR^n$, one asks whether it is possible to extract a subset of size $\ord{n}$ that, after potential reweighting, still constitutes a frame. This problem is closely related to Weaver's $\mathrm{KS}_2$ conjecture, see \cite[Thm.\ 2]{weaver2004kadison} or \cite[Conj.\ 1.2]{marcus2015interlacing}. Weaver's conjecture (now theorem) is equivalent to the Kadison--Singer problem, and forms the basis of the proof in \cite{marcus2015interlacing} of the latter.
The connection to sampling for least-squares approximation comes from the Marcinkiewicz--Zygmund inequality \ef{MZ-inequality}, which can be recast as a frame condition for the vectors $x^{(1)},\ldots,x^{(n)}$ defined by $x^{(i)} = ( \sqrt{w(x_i)} \phi_j(x_i)  )^{n}_{j=1}$, $i = 1,\ldots,n$.

The work \cite{marcus2015interlacing} (as well as extensions due to \cite{nitzan2016exponential}) has been used to show the existence of sampling points that result in $\ord{n}$ sample complexity for a fixed, but arbitrary subspace $\cP$. See \cite{limonova2022sampling,temlyakov2021optimal,nagel2021new,krieg2021function,krieg2021functionII,dolbeault2022optimal,dolbeault2023sharp} and references therein. Unfortunately, these works are impractical, as the computational cost of constructing the subsample is at least exponential in $n$.
Fortunately, recent progress has been made by using the approach of \cite{batson2014twice}, as well as techniques from \cite{lee2018constructing}, leading to practical algorithms that run in polynomial time. See \cite{bartel2023constructive} and \cite{dolbeault2024randomized} for two such approaches, as well as \cite{chen2019active} for related work in the discrete setting. A significant result of \cite{dolbeault2024randomized} establishes polynomial-time algorithms that also work down to the interpolation regime $m = n = \dim(\cP)$, albeit with constants in the error bounds that grow algebraically with $n$.

\subsubsection*{Sampling numbers and information-based complexity}

Another major area of focus in the last several years has been the use of (weighted) least-squares approximation, Christoffel sampling and subsampling techniques to provide new results in \textit{information-based complexity} \cite{novak2010trac,novak2008trac}. In this line of work, one considers a compact subset $\cK$ of a Banach space, then studies objects such as the \textit{(linear) sampling numbers} for $\cK$. These measure how well, in a worst-case sense, one can approximate functions in $\cK$ from $m$ arbitrary pointwise samples using arbitrary (linear) reconstruction maps. New results have emerged that bound the sampling numbers of an arbitrary $\cK$ in terms of its \textit{Kolmogorov $n$-width}, i.e., the best approximation error that can be achieved uniformly over $\cK$ using any $n$-dimensional subspace. These results show that pointwise samples (known as \textit{standard information}) can constitute near-optimal information for recovery. Some of the core ideas of this work can be found in Theorem \ref{thm:ullrich-approach}, including the construction of the measure \ef{opt-meas-krieg}, which is due to  \cite{krieg2021function,krieg2021functionII}.
Note that sampling numbers are often formulated with respect to the $L^2$-norm (as in this article), but recent works also consider other $L^p$-norms -- in particular, the uniform norm. For a selection of the many recent results in this direction, see \cite{bartel2023constructive,dolbeault2023sharp,krieg2023sampling,dolbeault2022optimal,temlyakov2021optimal,limonova2022sampling,nagel2021new,krieg2021function,krieg2021functionII} and references therein.

\subsubsection*{Sampling discretizations}

In tandem with these efforts, there has also been a focus on the development and systematic study of sampling discretizations using these ideas, both in the $L^2$-norm such as in \ef{MZ-inequality} and in other $L^p$-norms. We refer to \cite{kashin2022sampling,temlyakov2018marcinkiewicz} for reviews, as well as \cite{grochenig2020samplinng,dai2021entropy,dai2021sampling,limonova2022sampling}, and references therein. Note that $L^{\infty}$-norm sampling discretizations are related to the construction of \textit{weakly admissible meshes} \cite{xu2023randomized}. See also \cite{krieg2023sampling} for recent work which gives an essentially tight characterization.

As we have seen, sampling discretization are sufficient for accurate and stable recovery via (weighted) least squares. However, they are also necessary conditions for stable recovery by \textit{any} method. Modifying \cite[Rem.\ 6.2]{adcock2024learning}, let $\cR : \bbC^m \rightarrow L^2_{\varrho}(D)$ be an arbitrary reconstruction map and suppose that $\cR$ is \textit{$\delta$-accurate} over $\cP$, i.e.,
\be{
\label{R-accurate}
\nm{p - \cR( (p(x_i) )^{m}_{i=1})}_{L^2_{\varrho}(D)} \leq \delta \nm{p}_{L^2_{\varrho}(D)},\quad \forall p \in \cP.
}
Note that this holds with $\delta = 0$ for weighted least-squares whenever $\alpha_w > 0$, due to Lemma \ref{l:wLS-acc-stab}. Now let $\cF$ be any set of functions that are defined at the sample points $x_1,\ldots,x_m$ and suppose that $\cP \subseteq \cF$. Then it is a short argument to show that the \textit{$\epsilon$-Lipschitz constant}
\bes{
L_{\epsilon}(\cR ; \cF) = \sup_{f \in \cF} \sup_{0 < \nm{e}_{2,w} \leq \epsilon} \frac{\nm{\cR((f(x_i))^{m}_{i=1}) + e) - \cR(( f(x_i) )^{m}_{i=1}) }_{L^2_{\varrho}(D)}}{\nm{e}_{2,w}}
}
of the map $\cR$ satisfies
\bes{
L_{\epsilon}(\cR ; \cF)  \geq (1-\delta)/\alpha_w,\quad \forall \epsilon > 0,
}
where $\alpha^2_w$ is the lower constant in the sampling discretization \ef{MZ-inequality}. It follows that a reconstruction map cannot be both accurate (even over $\cP$, as in \ef{R-accurate}) \textit{and} stable without a sampling discretization.

\subsubsection*{Alternative sampling methods}

Many other sampling methods have been proposed over the last decade, especially in the context of high-dimensional polynomial approximation. However, these generally lack near-optimal sample complexity guarantees. See \cite[\S 8.1.1]{adcock2022sparse} and \cite{hadigol2018least} and references therein for overviews.

A limitation of Christoffel sampling is that i.i.d.\ points may cluster, thereby reducing the practical efficiency of the scheme. Most recently, a number of works have explored ideas such as \textit{volume sampling} \cite{deshpande2006volume} using \textit{determinantal point processes} to overcome this limitation. These are well-known concepts in machine learning \cite{kulesza2012determinantal,derezinski2018leveraged}, in which non-independent samples are drawn from a measure that promotes repulsion between the points. This transpires to be closely related to Christoffel sampling, since the marginals of the sample points follow the same distribution. The application of volume sampling to least-squares approximation in arbitrary subspaces has been considered in \cite{belhadji2023signal,trunscke2024optimal} for reproducing kernel Hilbert spaces and \cite{nouy2025weighted} for general spaces, along with its theoretical analysis and comparison with Christoffel sampling. Despite practical benefits, however, it is as of yet unclear whether this offers theoretical advantages over Christoffel sampling \cite{nouy2025weighted}.

\subsubsection*{Quadrature-based alternatives to least squares}

In the hierarchical setting (Remark \ref{rem:hierarchical}) it is possible to use quadrature-based methods instead of least-squares approximation \cite{krieg2019optimal,wasilkowski2007power}. Suppose that $\cP^{(k)} = \spn \{ \phi_{1},\ldots,\phi_{n_k} \}$, $k \in \bbN$, for some orthonormal basis $\{ \phi_i \}^{\infty}_{i=1}$. An approximation $\hat{f}^{(1)} \in \cP^{(1)}$ is first constructed by approximating the coefficients $c_i = \ip{f}{\phi_i}_{L^2_{\varrho}(D)}$, $i = 1,\ldots,n_1$, using Monte Carlo quadrature with $m_1$ points drawn i.i.d. from the Christoffel sampling measure associated to $\cP^{(1)}$. Next, at step $k \geq 1$ one draws $m_k$ new quadrature points i.i.d. from the Christoffel sampling measure associated to $\cP^{(k)}$ and updates the current approximation using a Monte Carlo quadrature approximation to the projection of the residual $f - \hat{f}^{(k-1)}$ onto $\cP^{(k)}$. This approach has several desirable properties. First, it is computational simpler than least-squares approximation, since it does not require solving an algebraic least-squares problem. Second, under certain assumptions, it requires only $\ord{n_k}$ samples to compute the $k$th approximation, rather than the $\ord{n_k \log(n_k)}$ samples required by the latter. Specifically, if the approximation errors $e_n(f)$ in \ef{e-def} are \textit{regularly decreasing} -- for example, algebraically decreasing -- then this method yields a nonuniform bound in expectation with the same rate of decrease with respect the number of samples $m$, up to constants \cite{krieg2019optimal}.

\subsubsection*{The effect of numerical redundancy}

In some problems, $\cP$ is described by a spanning set $\{ \psi_i \}^{n}_{i=1}$ that may be (numerically or analytically) redundant. This arises, for instance, when $\cP$ is spanned by the first $n$ elements of a \textit{frame} for $L^2_{\varrho}(D)$ \cite{adcock2019frames}. Such redundancy has a regularizing effect, which may significantly change the sampling question \cite{adcock2020frames,herremans2025sampling}. For example, uniform sampling with certain frames of polynomials has near-optimal sample complexity \cite{adcock2023fast}, whereas, as discussed in \S \ref{ss:MC-sampling}, with orthonormal bases of polynomials it is highly suboptimal. Such regularization also changes the optimal (Christoffel) sampling measure \cite{herremans2025sampling}, which now becomes a continuum analogue of so-called \textit{ridge} leverage scores \cite{alaoui2015fast}. See \cite{herremans2025sampling} for an in-depth analysis of sampling in the setting of numerical redundancy.

\subsubsection*{Adaptive methods}\label{ss:adaptive-approx}

Finally, we briefly mention the prospect of adaptive methods. While these methods typically lack full theoretical guarantees, they can prove extremely effective in practice. In a variation of Remark \ref{rem:hierarchical}, in an adaptive scheme one also chooses each subspace $\cP^{(k)}$ adaptively based on the previous approximation $\hat{f}^{(k-1)}$. In this case, we term this procedure an \textit{adaptive} approximation scheme. If given a dictionary of candidate basis functions to use to build the spaces $\cP^{(k)}$, this can be done using greedy methods \cite{temlyakov2011greedy}, as in \cite{migliorati2015adaptive,migliorati2019adaptive} (which are themselves based on adaptive quadrature routines \cite{gerstner2003dimension}). Moreover, adaptive methods can also be used when constructing approximations in complex, nonlinear approximation spaces. See \S \ref{s:conclusion} for some further discussion.

\section{Beyond linear spaces and pointwise samples}\label{s:general-framework}

Up to now, we have considered approximating an unknown function $f : D \rightarrow \bbC$ from a collection of $m$ pointwise samples in an $n$-dimensional subspace $\cP \subseteq L^2_{\varrho}(D)$. In this final section, we introduce a general framework that significantly extends this setup. This section is primarily based on the framework introduced in \cite{adcock2023cs4ml}, which was then further extended in \cite{adcock2024unified}. Unlike in previous sections, our presentation will now be less thorough: we aim to convey the main ideas without the full details or variations. See \cite{adcock2023cs4ml,adcock2024unified} for in-depth treatments, and \cite{eigel2022convergence,trunschke2022convergence} for related work.

There are four main extensions we now address:
\enum{
\item[(i)] The target object $f$ need not be a scalar-valued function, but simply an element of an abstract Hilbert space $\bbX$. 
\item[(ii)] The measurements arise as evaluations of arbitrary linear operators, which may, for instance, be scalar-, vector- or function space-valued. 
\item[(iii)] There may be $C \geq 1$ distinct processes generating the measurements. 
\item[(iv)] The approximation space $\cP$ need not be a finite-dimensional subspace of $\bbX$. 
}
Examples that motivate these generalizations are discussed in \S \ref{ss:framework-examples}.

\subsection{The general framework}\label{ss:general-setup}

Let $(\Omega,\cF,\bbP)$ be a probability space, $\bbX$ be a separable Hilbert space and consider a normed vector subspace of $\bbX_0 \subseteq \bbX$, termed the \textit{object space}. Let $C \geq 1$ be the number of measurement processes. For each $c = 1,\ldots,C$, let $(D_c,\cA_c,\varrho_c)$ be a probability space, which we term the \textit{measurement domain}, $\bbY_c$ be a Hilbert space, which we term the \textit{measurement space}, and
\bes{
L_c : D_c \rightarrow \cB(\bbX_0,\bbY_c)
}
be a mapping from $D_c$ to the space of $\cB(\bbX_0,\bbY_c)$ of bounded linear operators $\bbX_0 \rightarrow \bbY_c$, which we term the \textit{sampling operator}.

 For each $c$, let $\mu_{c}$ be such that $(D_c,\cA_c,\mu_{c})$ is a probability space. We assume that $\mu_c$ is absolutely continuous with respect to $\varrho_c$ and its Radon--Nikodym derivative $\nu_c : D_c \rightarrow \bbR$ is strictly positive almost everywhere on $\mathrm{supp}(\varrho_c)$. By definition
\bes{
\int_{D_c} \nu_c(\theta) \D \varrho_c(\theta) = 1.
}
Note that, in what follows, we generally use $\theta$ or $\theta_c$ to denote the variable in $D_c$, rather than $x$. For convenience, we also define the weight function $w_c : D_c \rightarrow \bbR$ by $w_c(\theta) = 1/\nu_c(\theta)$, $\theta \in D_c$.

Let $m_1,\ldots,m_C \in \bbN$, where $m_c$ is the number of measurements in the $c$th measurement process.  
We now draw samples independently with $\theta_{ic} \sim \mu_{c}$, $i = 1,\ldots,m_c$, $c = 1,\ldots,C$ and consider the measurements
\be{
\label{general-meas}
y_{ic} = L_c(\theta_{ic})(f) + e_{ic} \in \bbY_c,\quad i =1,\ldots,m_c,\ c = 1,\ldots,C,
}
where $e_{ic} \in \bbY_c$ is a noise term.
Finally, we let $\cP \subseteq \bbX_0$ be the \textit{approximation space} (which now need not be a linear space) and consider the empirical least-squares fit
\be{
\label{general-least-squares}
\hat{f} \in \argmin{p \in \cP} \sum^{C}_{c=1} \frac{1}{m_c} \sum^{m_c}_{i=1} w_c(\theta_{ic}) \nm{y_{ic} - L_c(\theta_{ic})(p) }^2_{\bbY_c}.
}
Note that computing solutions to \ef{general-least-squares} may be very challenging when $\cP$ is a nonlinear set. However, this is highly dependent on the choice of $\cP$, and therefore not the focus of this article.

\subsection{Examples}\label{ss:framework-examples}

As discussed in \cite{adcock2023cs4ml,adcock2024unified}, this framework includes many problems of practical relevance. We now summarize several examples. We start by showing that it generalizes the setup of \S \ref{s:prelims}.

\pbk
\textit{(i) Scalar-valued function approximation from pointwise samples}.
The can be formulated as follows. Let $C = 1$, $\bbX = L^2_{\varrho}(D)$, $\bbX_0 = C(\overline{D})$, $D_c = D$, $\varrho_1 = \varrho$ and $\bbY_1 = \bbR$ (with the Euclidean inner product), and define  $L_1 = L$ as the pointwise sampling operator $L(x)(f) = f(x)$ for $x \in D$, $f \in \bbX_0$. Note that the measurements \ef{general-meas} and least-squares fit \ef{general-least-squares} reduce to \ef{f_meas} and \ef{wls-prob}, respectively

\pbk
\textit{(ii) Function approximation from gradient-augmented samples.}
A simple modification of (i) involves sampling both the function and its gradient. This arises in various applications, including parametric PDEs and UQ \cite{alekseev2011estimation,peng2016polynomial,oleary-roseberry2022derivative-informed,oleary-roseberry2022derivative-informed-2}, seismology, Physics-Informed Neural Networks (PINNs) for PDEs \cite{feng2022gradient-enhanced,yu2022gradient-enhanced} and various other deep learning settings \cite{czarnecki2017sobolev}. Suppose that $D \subseteq \bbR^d$. Then this problem can be cast into the general framework by letting $C = 1$, $\bbX = H^1_{\varrho}(D)$ be the Sobolev space of order one, $\bbX_0 = C^1(\overline{D})$, $\bbY_1 = \bbR^{d+1}$ and $L_1 = L$ be defined by $L(x)(f) = (f(x) , \nabla f(x)^{\top} )^{\top}$.

The main difference between this and (i) is that the samples are vector-valued. Further generalizations are also possible. For example, in some cases it may be too expensive to evaluate the gradient at every point. Let $m_1$ be the number of function samples and $m_2$ be the number of function-plus-gradient samples. As shown in \cite[Sec.\ B.7]{adcock2023cs4ml}, we can consider this as a multimodal sampling problem with $C = 2$, $\bbY_1 = \bbR$, $\bbY_2 = \bbR^{d+1}$ and sampling operators $L_1(x)(f) = f(x)$ and $L_2(x)(f) = (f(x) , \nabla f(x)^{\top} )^{\top}$.

\pbk
\textit{(iii) Hilbert-valued function approximation from pointwise samples.} In some applications, the unknown is a function taking values in a Hilbert space $\bbV$. This arises in, for instance, parametric PDEs and UQ, where $f$ is the parametric solution map of a PDE whose (weak) solutions take values in $\bbV$. Approximating such a function from pointwise samples is easily considered in this framework. We use the setup of the scalar-valued case described above, except with $\bbY_1 = \bbR$ replaced by $\bbY_1 = \bbV$. One can also consider gradient-augmented samples, much as in the previous example. A further extension to this problem, which also fits within this framework, is that of operator learning \cite{boulle2024mathematical,kovachki2024operator}, in which the unknown is an operator between two Hilbert spaces.

\pbk
\textit{(iv) Image reconstruction.} We now briefly describe a seemingly quite different problem, based on \cite[Sec.\ C]{adcock2023cs4ml}. This example highlights that the general framework can handle both discrete and continuous settings, and measurements that do not arise as pointwise samples.
Consider a discrete $d$-dimensional image of size $n \times \cdots \times n$, which we may vectorize and express as a vector $f \in \bbC^N$, where $N = n^d$. Let $F \in \bbC^{N \times N}$ be the matrix of the $d$-dimensional discrete Fourier transform. In \textit{Fourier imaging} \cite{adcock2021compressive} the goal is to recover $f$ from a subset of its frequencies. If $\Omega \subseteq \{1,\ldots,N\}$, $| \Omega | = m$, is the set of frequencies sampled, then the measurements of $f$ are
\bes{
P_{\Omega} F f + e \in \bbC^m,
}
where $e \in \bbC^m$ is noise and $P_{\Omega} \in \bbC^{m \times N}$ is a matrix that selects the rows of $F$ corresponding to the indices in $\Omega$.
Fourier imaging arises in various applications, including Magnetic Resonance Imaging (MRI), Nuclear Magnetic Resonance (NMR) and radio interferometry \cite{adcock2021compressive}. A key question is how to best choose $\Omega$.
As described in \cite[Sec.\ C]{adcock2023cs4ml}, this problem can be cast into the general framework. The framework can also handle various practical constraints -- for instance, the fact that MRI devices cannot sample individual frequencies, but may only sample along piecewise smooth curves in frequency space, which leads to vector-valued measurements. The framework can also handle the more advanced scenario of \textit{parallel MRI}, where $C \geq 1$ coils simultaneously acquire measurements.

\pbk
\textit{(v) Other examples.} Various other families of problems can be considered within this framework. For instance, many standard measurement constructions in compressed sensing \cite{foucart2013mathematical} become special cases of this approach \cite{adcock2024unified}. One can also readily consider related problems such as matrix completion and, more generally, matrix recovery for linear measurements \cite{davenport2016overview}. Several other problems in sampling theory and signal processing also fit into this framework, such as \textit{mobile sampling} \cite{grochenig2015minimal}. This framework can also incorporate recovery problems involving function averages \cite{bruno2024polynomial}, as well as techniques such as \textit{stratification} and \textit{antithetics}, which are common variance reduction techniques in Monte Carlo integration \cite{owen2013monte}.

\pbk
\textit{(vi) Nonlinear approximation spaces.} Many recovery problems call for nonlinear approximation spaces. A standard example is the \textit{sparse regression} problem. Here, one typically considers the setup of \S \ref{s:prelims} with a linear subspace replaced by the set
\bes{
\cP = \left \{ \sum_{i \in S} c_i \phi_i : c_i \in \bbC,\ \forall i,\ S \subseteq \{1,\ldots,N\}, | S | = n \right \},
}
where $N \geq n \geq 1$ and $\{ \phi_i \}^{N}_{i=1} \subseteq L^2_{\varrho}(D)$ is some known dictionary of functions. The sparse regression problem has been studied extensively (see \cite{adcock2022towards} and references therein), especially in the context of dictionaries of polynomials, where it is termed \textit{sparse polynomial approximation} \cite{adcock2022sparse}. 

This is just one example of a nonlinear approximation space. There are many others. A partial list includes various `structured' sparse models, such as joint, group or block sparsity or sparsity in levels \cite{baraniuk2010model,bourrier2014fundamental,duarte2011structured,traonmilin2018stable}, low-rank matrix or tensor models \cite{davenport2016overview}, single- \cite{gajjar2023active} or multi-layer neural networks, tensor networks \cite{trunschke2021convergence,eigel2022convergence}, rational functions \cite{nakatsukasa2018aaa,herremans2023resolution,huybrechs2024sigmoid}, Fourier sparse functions \cite{erdelyi2020fourier} and spaces defined by generative models \cite{bora2017compressed}.

\subsection{The generalized Christoffel function}

The key tool in our analysis is a generalization of the Christoffel function (Definition \ref{def:christoffel}).

\defn{
[Generalized Christoffel function]
Let $\bbX$ and $\bbX_0$ be as above, $\bbY$ be a Hilbert space, $(D,\cA,\varrho)$ be a measure space, $L : D \rightarrow \cB(\bbX_0,\bbY)$ and $\cP \subseteq \bbX_0$, $\cP \neq \{ 0 \}$. The \textit{generalized Christoffel function of $\cP$ with respect to $L$} is the function $\cK = \cK(P,L) : D \rightarrow \bbR \cup \{ \infty \}$ defined by
\bes{
\cK(\theta) = \cK(\cP,L)(\theta) = \sup \left \{ \frac{\nm{L(\theta)(f)}^2_{\bbY}}{\nm{f}^2_{\bbX}} : f \in \cP,\ f \neq 0 \right \},\quad \forall \theta \in D.
}
}
Notice that $\cK$ reduces to the standard Christoffel function \ef{Kappa-def} in the case of (i). In general, $\cK$ measures how large the measurement $L(\theta)(f)$ of an arbitrary $f \in \cP$ can be (in norm) at an index $\theta \in D$ in relation to the norm of $f$. For instance, in the Fourier imaging problem (iv) it measures how large the Fourier transform can be at a given frequency for an element of $\cP$ in relation to its norm. We remark in passing that $\cK$ inherits some of the properties of the standard Christoffel function. See, e.g., \cite[Lem.\ E.1]{adcock2023cs4ml}.

Much like in \ef{kappa-w-def}, given a nonnegative weight function $w : D \rightarrow \bbR$, we also define
\be{
\label{kappa-w-def-general}
\kappa_{w} = \kappa_{w}(\cP , L) = \esssup_{\theta \sim \varrho} w(\theta) \cK(\cP,L)(\theta).
}

\subsection{Theoretical guarantee}

We now present a theoretical guarantee for this framework. We first require several assumptions.

\begin{assumption}[Nondegeneracy of the sampling operators]
\label{ass:nondegeneracy}
For each $c = 1,\ldots,C$ and $f \in \bbX_0$, the map $ \theta \in D_c \mapsto L_c(\theta)(f) \in \bbY_c$ is measurable and the function $ \theta \in D_c \mapsto \nm{L_c(\theta)(f)}^2_{\bbY_c} \in \bbR$ is integrable. Moreover, there are constants $0 < a \leq b <\infty$ such that
\be{
\label{nondegeneracy}
a \nm{f}_{\bbX} \leq \sqrt{\sum^{C}_{c=1} \int_{D_c} \nm{L_c(\theta_c)(f)}^2_{\bbY_c} \D \varrho_c(\theta_c)} \leq b \nm{f}_{\bbX},\quad \forall f \in \bbX_0.
}
\end{assumption}
We remark in passing that the lower bound \ef{nondegeneracy} is, in fact, only required to hold for $f \in \cP'$, where $\cP' = \cP - \cP = \{ p_1 - p_2 : p_1,p_2 \in \cP \}$ is the difference set (see the proof of Theorem \ref{thm:gen-framework} below, as well as that of \cite[Thm.\ E.2]{adcock2024unified}). 
Assumption \ref{ass:nondegeneracy} says that the action of the sampling operators preserves the norm of any $f \in \bbX_0$, up to constants. Note that it holds trivially with $a = b = 1$ in the standard problem (i), since in that case we have $C = 1$ and
\bes{
\int_{D_1} \nm{L_1(\theta_1)(f)}^2_{\bbY_1} \D \varrho_1(\theta_1) = \int_{D} |f(x)|^2 \D \varrho(x) = \nm{f}^2_{L^2_{\varrho}(D)} = \nm{f}^2_{\bbX}.
}
All other examples discussed in \S \ref{ss:framework-examples} can also be formulated so that this assumption also holds. 

Recall that in (i), stability and accuracy are ensured by the sampling discretization \ef{samp-disc}. The middle term in this inequality is an empirical approximation to the $L^2_{\varrho}$-norm. An analogous concept arises in the analysis of this general setting. Given $\{ \theta_{ic} : i = 1,\ldots,m_c,\ c = 1,\ldots,C\}$, we say that \textit{empirical nondegeneracy} holds for $\cP$ with constants $0 < \alpha_w \leq \beta_w < \infty$ if
\be{
\label{emp-nondegen}
\alpha_w \nm{q}_{\bbX} \leq \sqrt{\sum^{C}_{c=1} \frac{1}{m_c} \sum^{m_c}_{c=1} w(\theta_{ic}) \nm{L_c(\theta_{ic})(q)}^2_{\bbY_c} } \leq \beta_w \nm{q}_{\bbX},\quad \forall q \in \cP' = \cP - \cP.
}
This can be seen as a generalization of the well-known \textit{Restricted Isometry Property (RIP)} in compressed sensing \cite{foucart2013mathematical}. In the case of sparse regression (see (vi) above), it is sometimes termed a \textit{universal} sampling discretization \cite{dai2023universal}.

\begin{assumption}[Union-of-subspaces model]
\label{ass:UofS}
The set $\cP' = \cP - \cP$ satisfies the following.
\enum{
\item[(a)] $\cP'$ is a cone, i.e., $t p \in \cP'$ for any $t \geq 0$ and $p \in \cP'$.
\item[(b)] $\cP' \subseteq \cQ_1 \cup \cdots \cup \cQ_d = : \cQ$, where each $\cQ_i \subseteq \bbX_0$ is a subspace of dimension $n$.
}
\end{assumption}
This trivially holds with $d = 1$ and $\cQ_1 = \cP' = \cP$ when $\cP$ is an $n$-dimensional subspace. In general, Assumption \ref{ass:UofS} is an extension of the \textit{union-of-subspaces} model, which is well-known in compressed sensing \cite{baraniuk2010model,duarte2011structured}. It includes many nonlinear model classes used in practice, such as sparse regression and its various generalizations (see (vi) above).

For succinctness, we now only consider error bounds in expectation (bounds in probability could also be obtained). As in \S \ref{ss:err-bds-exp}, we introduce the truncation operator $\cT_{\sigma} : \bbX \rightarrow \bbX, g \mapsto \min  \{ 1 , \sigma / \nm{g}_{\bbX} \} g$, where $\sigma \geq 0$ is a constant. Given a minimizer $\hat{f}$ of \ef{general-least-squares}, we define $\hat{f}^{\mathsf{te}} = \cT_\sigma(\hat{f})$. Note that we do not consider the other estimator $\hat{f}^{\mathsf{ce}}$ introduced in \S \ref{ss:err-bds-exp}, although it could be readily formulated in this setting. The reason is that when $\cP$ is a nonlinear space, it is generally not possible to certify (in polynomial time) that \ef{emp-nondegen} holds. In sparse regression, for instance, this is equivalent to certifying that a given matrix has the RIP -- a well-known NP-hard problem.

\thm{
\label{thm:gen-framework}
Consider the setup of \S \ref{ss:general-setup} and suppose that Assumptions \ref{ass:nondegeneracy} and \ref{ass:UofS} hold. Let $0 < \epsilon < 1$, $\kappa_{w_c}$ be as in \ef{kappa-w-def-general} and suppose that
\be{
\label{mc-cond-1}
m_c \gtrsim a^{-2} \cdot \kappa_{w_c}(\cP'' , L_c) \cdot \left ( \log(2d/\epsilon) + n \right ), \quad c = 1,\ldots,C,
}
where $\cP'' = \cP' - \cP'$, or, if $\cQ$ is as in Assumption \ref{ass:UofS}(b), 
\be{
\label{mc-cond-2}
m_c \gtrsim a^{-2} \cdot \kappa_{w_c}(\cQ , L_c) \cdot \log(2nd/\epsilon),  \quad c = 1,\ldots,C.
}
Then, for any $f \in \bbX_0$, $\sigma \geq \nm{f}_{\bbX}$ and noise $\{ e_{ic} \}$, the estimator $\hat{f}^{\mathsf{tr}}$ satisfies 
\bes{
\bbE \left[ \nm{f - \hat{f}^{\mathsf{tr}}}^2_{\bbX} \right ] \lesssim \frac{b^2}{a^2} \cdot \inf_{p \in \cP} \nm{f - p}^2_{\bbX} + \frac{\nm{e}^2_2 }{ a^2}  + \sigma^2 \epsilon,
}
where $\nm{e}^2_{2} = \sum^{C}_{c=1} \frac{1}{m_c} \sum^{m_c}_{i=1} \nm{e_{ic}}^2_{\bbY_c}$.
}

Note that \ef{mc-cond-1} involves $\kappa_{w_c}(\cdot , L_c)$ evaluated over $\cP'' = \cP' - \cP'$. One can replace this with just $\cP'$, at the cost of a more complicated log term (see the proof of Theorem \ref{thm:gen-framework} and \cite[Thm.\ E.2]{adcock2024unified}).

\subsection{Christoffel sampling}

Much as in \S \ref{s:near-opt}, we can use Theorem \ref{thm:gen-framework} to optimize the sampling measures $\mu_c$. Let $\cU = \cP'' $ in the case of \ef{mc-cond-1} or $\cU = \cQ$ in the case of \ef{mc-cond-2}. Then, using \ef{kappa-w-def-general}, we now choose
\bes{
w^{\star}_c(\theta) = \left ( \frac12 + \frac12 \frac{\cK(\cU , L_c)(\theta) }{\int_{D_c} \cK(\cU , L_c)(\theta) \D \varrho_c(\theta) } \right )^{-1},
}
which gives the sampling measures
\bes{
\D \mu^{\star}_c(\theta) = \left ( \frac12 + \frac12 \frac{\cK(\cU , L_c)(\theta) }{\int_{D_c} \cK(\cU , L_c)(\theta) \D \varrho_c(\theta) } \right ) \D \varrho_c(\theta),\quad c = 1,\ldots,C.
}
We term this \textit{Christoffel sampling}.\footnote{Note we may assume without loss of generality that $\int_{D_c} \cK(\cU , L_c)(\theta) \D \varrho_c(\theta)  > 0$. If not, the sampling operator $L_c$ simply yields zero measurements over the space $\cU$ almost everywhere, and can therefore be excluded. Nondegeneracy \ef{nondegeneracy} implies that there is at least one sampling operator yielding nonzero measurements over $\cU$.}
Substituting this into \ef{mc-cond-1} yields the measurement conditions
\be{
\label{mc-cond-1-opt}
m_c \gtrsim a^{-2} \cdot \left ( \int_{D_c} \cK(\cP'' , L_c)(\theta) \D \varrho_c(\theta) \right ) \cdot (\log(2d/\epsilon) + n) ,\quad c = 1,\ldots,C,
}
or, in the case of \ef{mc-cond-2},
\be{
\label{mc-cond-2-opt}
m_c \gtrsim a^{-2} \cdot\left ( \int_{D_c} \cK(\cQ, L_c)(\theta) \D \varrho_c(\theta) \right ) \cdot \log(2nd/\epsilon) ,\quad c = 1,\ldots,C.
}
This approach is `optimal' in the sense that it minimizes (up to a factor of $2$) the bound \ef{mc-cond-1} over all possible sampling measures $\mu_c$. When $\cP$ is an $n$-dimensional subspace -- in which case $\cP''  = \cP' = \cP$ and $\cQ$ in Assumption \ref{ass:UofS} can be chosen as $\cQ = \cP$ -- it is a short argument using the nondegeneracy condition \ef{nondegeneracy} to see that
\bes{
\sum^{C}_{c=1} \int_{D_c} \cK(\cP , L_c)(\theta) \D \varrho_c(\theta) \leq b n
}
(see \cite[Cor.\ 4.7]{adcock2023cs4ml}). Hence, if each $m_c$ is chosen proportional to the right-hand side in \ef{mc-cond-2-opt}, then the total number of measurements satisfies the near-optimal log-linear scaling
\bes{
m = m_1 + \cdots + m_C \lesssim (b/a)^2 \cdot n \cdot \log(2n/\epsilon).
}
Unfortunately, in the general case of a nonlinear set $\cP$, there is no clear way to relate the integral in \ef{mc-cond-1-opt} to explicit quantities such as $n$ and $d$. It is possible to show the bound
\bes{
m = m_1 + \cdots + m_C \lesssim (b/a)^2 \cdot n \cdot d \cdot \log(2n/\epsilon)
}
(see \cite[Cor.\ 4.7]{adcock2023cs4ml} once more),
where we recall that $d$ is the number of subspaces in Assumption \ref{ass:UofS}(b). For fixed and small $d$, this is near-optimal. However, in cases such as sparse regression, $d \gg  1$. Fortunately, a more refined analysis is possible in these cases. See \cite{adcock2022towards,adcock2023cs4ml,adcock2024unified} for discussion.

While it is difficult to provide explicit measurement conditions in the general case, it is possible to gain some insight over why Christoffel sampling, in general, improves over Monte Carlo sampling, i.e., the case where $\mu_c = \varrho_c$, $\forall c$. Since $w_c \equiv 1$ in this case, \ef{kappa-w-def-general} and  Theorem \ref{thm:gen-framework} provides the measurement conditions for Monte Carlo sampling of the form
\bes{
m_c \gtrsim a^{-2} \left ( \esssup_{\theta \sim \varrho_c} \cK(\cP''  , L_c)(\theta) \right )\cdot (\log(2d/\epsilon) + n ),\quad c = 1,\ldots,C,
}
and likewise in the case of \ef{mc-cond-2}.
Therefore, comparing with \ef{mc-cond-1-opt}, the improvement of Christoffel sampling, in general terms, can be equated to the difference between the supremum of the (generalized) Christoffel function and its integral (mean). In particular, if this function is sharply peaked, then we expect significant improvements, while if it is approximately flat, then we expect less improvement. Such observations are witnessed in numerical experiments \cite{adcock2023cs4ml,adcock2024unified}.

\subsection{Summary}

In summary, Christoffel functions and, consequently, Christoffel sampling extend to much more general types of recovery problems, where the unknown need not be a scalar-valued function, the samples need not be pointwise evaluations and the approximation space need not be a linear subspace. For linear subspaces this leads to near-optimal sample complexity bounds, as before. For nonlinear spaces, while it is unclear if Christoffel sampling is near-optimal, it always leads to better sample complexity bounds than those of Monte Carlo sampling.

\section{Conclusions and outlook}\label{s:conclusion}

In this article, we have surveyed recent advances in optimal random sampling, termed Christoffel sampling, which arises from establishing the key role that the Christoffel function plays in the sample complexity of least-squares approximation with random samples. We have also seen in \S \ref{s:general-framework} how these ideas naturally extend to more general types of measurements and play a key role in the sample complexity even for nonlinear spaces. We now offer some concluding thoughts.

First, although the picture for pointwise sampling in linear spaces is increasingly mature, there remain various open questions. While optimal (i.e., $\ord{n}$) sampling strategies that are practical (i.e., implementable in polynomial time) are now known (recall \S \ref{s:further}), future investigations are needed on their practical efficacy, especially in the interpolation regime. There are also open questions about uniform recovery, which are especially relevant to sampling numbers and questions in information-based complexity. Finally, the question of optimal sampling with hierarchical or adaptive schemes has not yet been addressed.

By contrast, nonlinear approximation spaces pose many more open problems. First, even in relatively well-studied settings such as sparse regression, it is unknown whether Christoffel sampling generally leads to near-optimal sample complexity bounds \cite{adcock2022towards}. Less is known about more complicated nonlinear spaces. See \cite{adcock2023cs4ml,adcock2024unified} for discussion. Second, there is also the practical matter of drawing samples from the resulting Christoffel sampling measure. This is very dependent on the particular nonlinear space and samples under consideration, and may be highly nontrivial. However, recent studies have constructed practical surrogate sampling measures in cases such as sparse regression with orthonormal bases \cite{adcock2022towards} and Fourier imaging with generative models \cite{adcock2023cs4ml}. Determining how to do this in other settings is an interesting topic for future work.

In the nonlinear setting, it is worth noting that Christoffel sampling is well-suited only when the approximation space $\cP$ has low intrinsic complexity that is comparable to the number of samples (e.g., $m \asymp n \log(n)$ in the case where $\cP$ is a linear space with $\dim(\cP) = n$). It is not well suited for `complex' approximation spaces, such as spaces of deep neural networks \cite{adcock2023cs4ml} or low-rank tensor networks \cite{eigel2022convergence}. Christoffel sampling can be implemented in an adaptive manner in such cases. Here, one alternates between adding samples and learning an approximation, and at each stage uses a linearization to obtain an intermediate linear approximation space over which Christoffel sampling can be performed. This idea was developed for approximating functions and solving PDEs via deep learning in \cite{adcock2022cas4dl} and \cite[Sec.\ D]{adcock2023cs4ml}, and later extended to more general approximation spaces in \cite{gruhlke2024optimal}. 

Finally, we remark that Christoffel sampling, in any of its guises, is not a panacea. Depending on the problem, there may be little or no benefit over standard Monte Carlo sampling. This is relevant in various applications, as Monte Carlo samples are commonly encountered in practice \cite{adcock2023monte}. It leads to another interesting line of research, which is understanding function classes where Monte Carlo sampling is near-optimal. One such case is
holomorphic function approximation in high dimensions. In a series of works \cite{adcock2023monte,adcock2024optimal,adcock2025optimal}, it has been shown that Monte Carlo sampling is near-optimal information for classes of infinite-dimensional holomorphic functions arising in parametric DE problems.  For another line of work in this direction involving Sobolev spaces, see \cite{krieg2024random,krieg2022recovery,krieg2025function}.

\section*{Acknowledgements}

The idea for this article came from a plenary talk given at the 2023 Foundations of Computational Mathematics conference. The author would like to thank the conference organizers, as well as the various participants for insightful comments and questions. Parts of this article were written during a visit to the Isaac Newton Institute for Mathematical Sciences, Cambridge for the programme \textit{Discretization and recovery in high-dimensional spaces}. The author would like to thank the institute for support and hospitality, and the programme participants for providing a stimulating environment. Finally, he would like to thank Simone Brugiapaglia, Matthew J. Colbrook, Daniel Fassler, Marta Ghirardelli, Astrid Herremans, Daan Huybrechs, David Krieg and Maksym Neyra--Nesterenko for helpful feedback.
This work was supported by EPSRC grant number EP/R014604/1 and NSERC grant number RGPIN-2021-611675.

{\small
\bibliographystyle{plain}
\bibliography{FoCMOptSampBib}
}

\appendix
	
\section{Proofs}\label{app:proofs}

In this appendix we give proofs for various results in the paper. We commence with Lemma \ref{l:wLS-acc-stab}, which is a standard result. We include a short proof for completeness.

\prf{
[Proof of Lemma \ref{l:wLS-acc-stab}]

Recall that $\alpha_{w} = \sigma_{\min}({A})$, where ${A}$ is the matrix defined in \eqref{ls-Ab}. This matrix is full rank since $m \geq n$ and $\alpha_{w} > 0$, and therefore the least-squares problem has a unique solution. 
Now let $p \in \cP$ be arbitrary and consider the variational form \ef{variational-form} applied with the element $\hat{f} - p \in \cP$. Subtracting $\ip{p}{\hat{f} - p}_{\mathsf{disc},w}$ from both sides, we get
\bes{
\nm{\hat{f} - p}^2_{\mathsf{disc},w} = \ip{f - p}{\hat{f} - p}_{\mathsf{disc},w} +  \frac1m \sum^{m}_{i=1} w(x_i) e_i \overline{\hat{f}(x_i) - p(x_i)},
}
and, by several applications of the Cauchy-Schwarz inequality, we deduce that
\be{
\label{disc-err-bd}
\nm{\hat{f} - p}_{\mathsf{disc},w} \leq \nm{f - p}_{\mathsf{disc},w} + \nms{e}_{2,w}. 
}
We now use the triangle inequality and \ef{MZ-inequality} to obtain
\bes{
\nm{f - \hat{f}}_{L^2_{\varrho}(\cU)} \leq \nm{f - p}_{L^2_{\varrho}(\cU)} + \nm{\hat{f} - p}_{L^2_{\varrho}(\cU)} \leq \nm{f - p}_{L^2_{\varrho}(\cU)} + \frac{1}{\alpha_w} \nm{\hat{f}-p}_{\mathsf{disc},w}.
}
The result now follows from \ef{disc-err-bd}.
}

We next prove Theorem \ref{t:alpha-beta-est}. This has now also become a standard exercise involving the \textit{matrix Chernoff bound} (see, e.g., \cite[Thm.\ 1.1]{tropp2012user-friendly}), which we repeat here for convenience. This bound was first used in the context of least-squares approximation from i.i.d.\ samples in \cite{cohen2013stability}.

\thm{
[Matrix Chernoff bound]
\label{t:matrix-chernoff}
Let ${X}_1,\ldots,{X}_{m}$ be independent, self-adjoint random matrices of dimension $n$.  Assume that 
\bes{
{X}_{i} \succeq {0}
\quad\text{and}\quad 
\lambda_{\max}({X}_i) \leq R
}
almost surely for each $i=1,\ldots,m$, and define
\bes{
\mu_{\min} = \lambda_{\min} \left ( \sum^{m}_{i=1} \bbE [{X}_i] \right ) \quad\text{and}\quad 
\mu_{\max} = \lambda_{\max} \left ( \sum^{m}_{i=1} \bbE[{X}_i] \right ).
}
Then, for $0 \leq \delta \leq 1$,
\bes{
\bbP \left ( \lambda_{\min}  \left ( \sum^{m}_{i=1} {X}_i \right ) \leq \left ( 1 - \delta \right ) \mu_{\min} \right ) \leq n \exp \left ( -\frac{\mu_{\min} ( (1-\delta) \log(1-\delta) + \delta )}{R} \right ),
}
and, for $\delta \geq 0$,
\bes{
\bbP \left ( \lambda_{\max}  \left ( \sum^{m}_{i=1} {X}_i \right ) \geq \left ( 1 + \delta \right ) \mu_{\max} \right ) \leq n \exp \left ( -\frac{\mu_{\max} ( (1+\delta) \log(1+\delta)-\delta)}{R} \right ).
}
}
Here and subsequently, we use the notation $X \succeq 0$ to mean that $X$ is nonnegative definite.

\prf{
[Proof of Theorem \ref{t:alpha-beta-est}]

Let $\{ \phi_i \}^{n}_{i=1}$ be an orthonormal basis for $\cP$ and $A$ be as in \ef{ls-Ab}. Then
\bes{
A^* A = \sum^{m}_{i=1} X_i,\qquad \text{where }X_i : =  \frac{1}{m} \left ( w(x_i) \overline{\phi_j(x_i)} \phi_k(x_i) \right )^{n}_{j,k=1} ,
}
is a sum of independent random matrices. Also, orthonormality of the basis functions implies that \ef{alpha-beta-sigma} holds, i.e., $\alpha^2_w = \lambda_{\min}(A^*A)$ and $\beta^2_w = \lambda_{\max}(A^*A)$.
We now wish to apply Theorem \ref{t:matrix-chernoff}. Due to \ef{mu_weight_fn}, the fact that $w = 1/\nu$ and orthonormality, we have
\bes{
(\bbE[A^*A])_{jk} = \sum^{m}_{i=1} (\bbE[X_i])_{jk} = \frac1m \sum^{m}_{i=1} \int_{D} w(x) \overline{\phi_j(x)} \phi_k(x) \D \mu_i(x) = \int_{D} \overline{\phi_j(x)} \phi_k(x) w(x) \nu(x) \D \varrho(x) = \delta_{jk}.
}
Hence $\bbE[A^*A] = I$ and therefore $\mu_{\min} = \mu_{\max} = 1$. Next, let $c \in \bbC^n$ be arbitrary. Then
\bes{
c^* X_i c = \frac{w(x_i)}{m}\left | \sum^{n}_{j=1} c_j \phi_j(x_i) \right |^2 = \frac{w(x_i)}{m} | p(x_i) |^2,\qquad \text{where }p = \sum^{n}_{j=1} c_j \phi_j.
}
We immediately deduce that $X_{i} \succeq 0$. Moreover, Parseval's identity implies that $\nms{c}^2_2 = \nms{p}^2_{L^2_{\varrho}(D)}$. Hence, using this and \ef{Kappa-def} and \ef{kappa-w-def}, we obtain 
\bes{
c^* X_i c \leq \frac{w(x_i)}{m} K(\cP)(x_i) \nms{p}^2_{L^2_{\varrho}(D)} \leq \frac{\kappa_w(\cP)}{m} \nms{c}^2_2.
}
Since $c$ was arbitrary and $X_i \succeq 0$, we conclude that
\bes{
\lambda_{\max}(X_i) \leq R : = \frac{\kappa_w(\cP)}{m}.
}
We are now ready to apply Theorem \ref{t:matrix-chernoff}. Using this and the union bound, we have
\eas{
\bbP \left ( \alpha_w \leq \sqrt{1-\delta}\text{ or }\beta_w \geq \sqrt{1+\delta} \right ) & \leq \bbP \left (\lambda_{\min}(A^*A) \leq 1-\delta \right ) + \bbP \left ( \lambda_{\max}(A^*A) \geq 1-\delta \right )
\\
& \leq n \exp \left ( - \frac{m a_{\delta}}{ \kappa_w(\cP)} \right) + n \exp \left ( - \frac{m b_{\delta}}{ \kappa_w(\cP)} \right ),
}
where $a_{\delta} =  ( (1-\delta) \log(1-\delta) + \delta )$ and $b_{\delta} = ( (1+\delta) \log(1+\delta)-\delta)$. Notice that $a_{\delta} \geq b_{\delta}$ and $b_{\delta} = 1/C_{\delta}$, where $C_{\delta}$ is as in \ef{m-bound-alpha-beta}. Therefore
\bes{
\bbP \left ( \alpha_w \leq \sqrt{1-\delta}\text{ or }\beta_w \geq \sqrt{1+\delta} \right ) \leq 2 n \exp \left ( -\frac{m}{C_{\delta} \kappa_w(\cP)} \right ) \leq \epsilon,
}
where in the last step we used \ef{m-bound-alpha-beta}. This completes the proof.
}

We next prove Corollary \ref{cor:err-prob-1}. For this, we require the following lemma, which is based on \cite{migliorati2019adaptive} (which is, in turn, based on \cite{cohen2013stability}).

\lem{
\label{lem:exp-inner-prod-perp}
Let $\cP \subset L^2_{\varrho}(D)$ be an $n$-dimensional subspace with orthonormal basis $\{ \phi_i \}^{n}_{i=1}$ and $\mu_{1},\ldots,\mu_m$ be probability measures satisfying Assumption \ref{ass:mixture}. Consider sample points drawn randomly and independently with $x_i \sim \mu_i$, $i = 1,\ldots,m$. Then
\bes{
\bbE \left[ \sum^{n}_{i=1} | \ip{g}{\phi_i}_{\mathsf{disc},w} |^2 \right ] \leq \frac{\kappa_w(\cP)}{m} \nms{g}^2_{L^2_{\varrho}(D)},\quad \forall g \in \cP^{\perp},
}
where  $w = 1/\nu$, $\ip{\cdot}{\cdot}_{\mathsf{disc},w}$ and $\kappa_w$ are as in \ef{mu_weight_fn},  \ef{semi-inner-product} and \ef{kappa-w-def}, respectively.
}
\prf{
Let $g \in \cP^{\perp}$ and $\l \in \{1,\ldots , n \}$ be arbitrary. Then
\eas{
\bbE \left [  |\ip{g}{\phi_l}_{\mathsf{disc},w}|^2 \right ] =& ~ \frac{1}{m^2} \sum^{m}_{i,j = 1} \bbE \left [ w(x_i) w(x_j) \overline{g(x_i)} g(x_j) \phi_l(x_i) \overline{\phi_l(x_j)} \right ]
\\
 = &~ \frac{1}{m^2}  \sum^{m}_{\substack{i,j = 1 \\ i \neq j}} \bbE \left [ w(x_i) g(x_i) \phi_l(x_i) \right ] \overline{\bbE \left [ w(x_j) g(x_j) \phi_l(x_j) \right ] }
 \\
&  + \frac{1}{m^2} \sum^{m}_{i=1} \bbE \left [ | w(x_i) g(x_i) \phi_l(x_i) |^2 \right ] .
}
Now \ef{mu_weight_fn} implies that
\bes{
\frac{1}{m} \sum^{m}_{i=1} \bbE \left [ w(x_i) g(x_i) \phi_l(x_i) \right ] = \frac{1}{m} \sum^{m}_{i=1} \int_{D} w(x) g(x) \phi_l(x) \D \mu_i(x) = \ip{g}{\phi_l}_{L^2_{\varrho}(D)} = 0,
}
since $g \in \cP^{\perp}$. Therefore
\eas{
\bbE \left [ |\ip{g}{\phi_l}_{\mathsf{disc},w}|^2 \right ] &=  \frac{1}{m^2} \sum^{m}_{i=1} \left ( \bbE \left [ | w(x_i) g(x_i) \phi_l(x_i) |^2 \right ] - \left | \bbE \left [ w(x_i) g(x_i) \phi_l(x_i) \right ] \right |^2 \right )
\\
& \leq \frac{1}{m^2} \sum^{m}_{i=1} \int_{D} (w(x))^2 | g(x) |^2 | \phi_l(x) |^2 \D \mu_i(x)
= \frac{1}{m} \int_{D} w(x) |g(x)|^2 | \phi_l(x) |^2 \D \varrho(x),
}
where in the final step we used \ef{mu_weight_fn} once more. 
We now apply \ef{Kappa-def-alt} and \ef{kappa-w-def} to obtain
\eas{
\sum^{n}_{l=1} \bbE \left [ |\ip{g}{\phi_l}_{\mathsf{disc},w}|^2 \right ] \leq \frac{1}{m} \int_{D} w(x) |g(x)|^2 \sum^{n}_{l=1} | \phi_l(x) |^2 \D \varrho(x)  \leq \frac{\kappa_w(\cP)}{m} \nms{g}^2_{L^2_{\varrho}(D)},
}
as required.
}

\prf{
[Proof of Corollary \ref{cor:err-prob-1}]

The condition on $m$ and Theorem \ref{t:alpha-beta-est} imply that \ef{alpha-beta-delta} holds with probability at least $1-\epsilon/2$. Therefore Lemma \ref{l:wLS-acc-stab} asserts that the weighted least-squares problem has a unique solution for any function that is defined at the sample points $x_i$ and any noise vector. Let $f \in L^2_{\varrho}(D)$. Then $f$ is defined at the $x_i$ with probability one. Now consider arbitrary noise $e = (e_i)^{m}_{i=1} \in \bbC^m$ and write $\hat{f}_e \in \cP$ for the corresponding weighted least-squares approximation from noisy samples $f(x_i) + e_i$. Let $p^* \in \cP$ be the (unique) element such that $\nm{f - p^*}_{L^2_{\varrho}(D)} = \inf_{p \in \cP} \nms{f - p}_{L^2_{\varrho}(D)}$ and write $g = f - p^*$. Let $\hat{g} = \hat{g}_0$ be the least-squares approximation to $g$ from noiseless samples $y_i = g(x_i)$ and $\hat{0}_e \in \cP$ be the least-squares approximation to the zero function from the noisy samples $y_i = e_i$. Notice that $\widehat{p^*} = \widehat{p^*}_0 = p^*$, as the weighted least-squares approximation is a projection in the discrete inner product. Since the least-squares approximation is also linear, we have $f - \hat{f}_e = f - \hat{f} - \hat{0}_e = g - \hat{g} - \hat{0}_e$.
This gives
\be{
\label{main-err-split}
\nm{f-\hat{f}_e}_{L^2_{\varrho}(D)} \leq  \nms{g}_{L^2_{\varrho}(D)} + \nms{\hat{g}}_{L^2_{\varrho}(D)} + \nms{\hat{0}_e}_{L^2_{\varrho}(D)}.
}
Lemma \ref{l:wLS-acc-stab} and the fact that \ef{alpha-beta-delta} holds imply that
\be{
\label{q-bd}
\nms{\hat{0}_e}_{L^2_{\varrho}(D)} \leq \frac{1}{\sqrt{1-\delta}} \nms{e}_{2,w}.
}
Now consider the term $\nms{\hat{g}}_{L^2_{\varrho}(D)} $ and write $\hat{g} = \sum^{n}_{i=1} \hat{c}_i \phi_i$ and $\hat{c} = (\hat{c}_i)^{n}_{i=1}$. The variational form \ef{variational-form} and the Cauchy--Schwarz inequality give
\bes{
\nms{\hat{g}}^2_{\mathsf{disc},w} = \ip{g}{\hat{g}}_{\mathsf{disc},w}= \sum^{n}_{j=1} \hat{c}_j \ip{g}{\phi_j}_{\mathsf{disc,w}} \leq \nms{\hat{c}}_2 \sqrt{\sum^{n}_{j=1} |\ip{g}{\phi_j}_{\mathsf{disc,w}}|^2 },
}
and \ef{MZ-inequality}, the fact that $\alpha_w \geq \sqrt{1-\delta}$ and Parseval's identity give
\bes{
\nm{\hat{g}}^2_{L^2_{\varrho}(D)} \leq \frac{1}{1-\delta} \nms{\hat{g}}^2_{\mathsf{disc},w} \leq \frac{1}{1-\delta} \nm{\hat{g}}_{L^2_{\varrho}(D)} \sqrt{\sum^{n}_{j=1} |\ip{g}{\phi_j}_{\mathsf{disc,w}}|^2 },
}
i.e.,
\be{
\label{g-bd-1}
\nm{\hat{g}}_{L^2_{\varrho}(D)} \leq \frac{1}{1-\delta} \sqrt{\sum^{n}_{j=1} |\ip{g}{\phi_j}_{\mathsf{disc,w}}|^2 }.
}
We now bound this term in probability. Consider the random variable $X = \sum^{n}_{j=1} |\ip{g}{\phi_j}_{\mathsf{disc,w}}|^2$. Since $p^*$ is the orthogonal projection of $f$ onto $\cP$, we have $g = f - p^* \in \cP^{\perp}$ and consequently Lemma \ref{lem:exp-inner-prod-perp} implies that
\bes{
\bbE[X] \leq \frac{\kappa_w(\cP)}{m} \nms{g}^2_{L^2_{\varrho}(D)}.
}
Hence, by Markov's inequality,
\bes{
\bbP \left ( \nm{\hat{g}}_{L^2_{\varrho}(D)} \geq \frac{1}{1-\delta} \sqrt{\frac{2 \kappa_{w}(\cP)}{m \epsilon}} \nms{g}_{L^2_{\varrho}(D)} \right ) \leq \bbP \left ( X \geq \frac{2 \bbE[X]}{\epsilon}  \right ) \leq \frac{\epsilon}{2}.
}
We deduce that 
\bes{
\nm{\hat{g}}_{L^2_{\varrho}(D)} \leq \frac{1}{1-\delta}  \sqrt{\frac{2 \kappa_{w}(\cP)}{m \epsilon}} \nms{g}_{L^2_{\varrho}(D)},
}
with probability at least $1-\epsilon/2$. Substituting this and \ef{q-bd} into \ef{main-err-split} we deduce, after an application of the union bound, that
\bes{
\nm{f-\hat{f}_e}_{L^2_{\varrho}(D)} \leq  \left ( 1 +  \frac{1}{1-\delta}  \sqrt{\frac{2 \kappa_{w}(\cP)}{m \epsilon}} \nms{g}_{L^2_{\varrho}(D)} \right ) \nms{g}_{L^2_{\varrho}(D)} + \frac{1}{\sqrt{1-\delta}} \nms{e}_{2,w},
}
with probability at least $1-\epsilon$. This completes the proof.
}

We next prove Corollary \ref{cor:err-prob-2}. This follows \cite[Lem.\ 7.11]{adcock2022sparse} and employs Bernstein's inequality.

\prf{
[Proof of Corollary \ref{cor:err-prob-2}]
Let $p = p^*$ be a polynomial attaining the infimum in \ef{ls-err-bd-prob-2}, $E$ be the event that \ef{alpha-beta-delta} holds and $F$ be the event that
\bes{
\nm{f - p}_{\mathsf{disc},w} \leq \sqrt{2} \left ( \nm{f - p}_{L^{2}_{\varrho}(D)} + \frac{\nm{\sqrt{w}(f-p)}_{L^{\infty}_{\varrho}(D)}}{\sqrt{k}}  \right ).
}
Suppose that $E$ and $F$ occur. Then Lemma \ref{l:wLS-acc-stab} implies that
\eas{
\nm{f - \hat{f}}_{L^2_{\varrho}(D)} & \leq \nm{f - p}_{L^2_{\varrho}(D)} + \frac{1}{\sqrt{1-\delta}} \nm{f - p}_{\mathsf{disc},w} + \frac{1}{\sqrt{1-\delta}} \nm{e}_{2,w}
\\
& \leq \left ( 1 + \sqrt{\frac{2}{1-\delta}} \right ) \nm{f - p}_{L^2_{\varrho}(D)} + \sqrt{\frac{2}{1-\delta}}  \frac{\nms{f-p}_{L^{\infty}_{\varrho}(D)}}{\sqrt{k}} + \frac{1}{\sqrt{1-\delta}} \nm{e}_{2,w}.
}
This yields \ef{ls-err-bd-prob-2}. Hence, by the union bound, it suffices to show that $\bbP(E^c),\bbP(F^c) \leq \epsilon /2$.

The fact that $\bbP(E^c) \leq \epsilon / 2$ follows immediately from the first condition on $m$ in \ef{m-conds-in-prob-2} and Theorem \ref{t:alpha-beta-est}. We now consider $\bbP(F^c)$. Define the random variables
\bes{
Z_i = w(x_i)| f(x_i) - p(x_i) |^2\quad \text{and}\quad X_i = Z_i - \bbE[Z_i].
}
Notice that $\frac1m \sum^{m}_{i=1} \bbE[Z_i] = \nms{f-p}^2_{L^2_{\varrho}(D)} : = a$
due to \ef{exp-sum-scaling}, and therefore
\bes{
\nm{f -p}^2_{\mathsf{disc},w} = \frac1m \sum^{m}_{i=1} Z_i = \frac1m \sum^{m}_{i=1} X_i + a.
}
The idea now is to use Bernstein's inequality to estimate the random variable $\sum^{m}_{i=1} X_i$. Let
\bes{
b = \esssup_{x \sim \varrho} w(x) | f(x) - p(x) |^2 \equiv \nm{\sqrt{w}(f - p)}^2_{L^{\infty}_{\varrho}(D)}
}
and notice that $X_i \leq Z_i \leq b$ and $-X_i \leq \bbE[Z_i] \leq b$
almost surely. Hence $|X_i| \leq b$ almost surely. We also have that $0 \leq Z_i \leq b$ almost surely, and therefore
\bes{
\sum^{m}_{i=1} \bbE[X_i^2] \leq \sum^{m}_{i=1} \bbE[Z_i^2] \leq b \sum^{m}_{i=1} \bbE[Z_i] = a b m.
}
Since $\bbE[X_i] = 0$, we may apply Bernstein's inequality (see, e.g., \cite[Cor.\ 7.31]{foucart2013mathematical}) to get
\bes{
\bbP \left ( \left | \frac1m \sum^{m}_{i=1} X_i \right | \geq t \right ) \leq 2 \exp \left ( -\frac{t^2 m/2}{a b + b t /3} \right ),\quad \forall t > 0.
}
Set $t = a + b/k$. Then it is a short argument involving the second condition in \ef{m-conds-in-prob-2} to show that $\frac{t^2 m/2}{a b + b t /3} \geq \log(4/\epsilon)$. Therefore,
\bes{
\nm{f - p}^2_{\mathsf{disc},w} \leq \left | \frac1m \sum^{m}_{i=1} X_i \right | + a <  2 a + b/k,
}
with probability at least $1-\epsilon/2$. Substituting the values for $a$, $b$ and using the inequality $\sqrt{s+t} \leq \sqrt{s} + \sqrt{t}$, we see that 
\bes{
\nm{f - p}_{\mathsf{disc},w} \leq \sqrt{2} \left ( \nm{f - p}_{L^2_{\varrho}(D)} + \frac{\nm{\sqrt{w}(f-p)}_{L^{\infty}_{\varrho}(D)}}{\sqrt{k}} \right ),
}
with probability at least $1-\epsilon/2$. Therefore, $\bbP(F^c) \leq \epsilon/2$, as required.
}

We now prove Lemma \ref{l:exp-bounds-chi} and Theorem \ref{t:err-bd-exp}. These ideas go back to \cite{cohen2013stability}, but the specific arguments are based on \cite{cohen2017optimal}.

\prf{
[Proof of Lemma \ref{l:exp-bounds-chi}]

The setup is the same as the proof of Corollary \ref{cor:err-prob-1}. However, since we now square the error terms, we use the fact that $g \in \cP^{\perp}$ to replace \ef{main-err-split} by
\bes{
\nm{f-\hat{f}_e}^2_{L^2_{\varrho}(D)} = \nms{g}^2_{L^2_{\varrho}(D)} + \nms{\hat{g} + \hat{0}_e}^2_{L^2_{\varrho}(D)} \leq  \nms{g}^2_{L^2_{\varrho}(D)} + 2 \nms{\hat{g}}^2_{L^2_{\varrho}(D)} + 2\nms{\hat{0}_e}^2_{L^2_{\varrho}(D)}.
}
Whenever $\nm{G - I}_{2} \leq \delta$ we have that \ef{alpha-beta-delta} holds, and therefore \ef{q-bd} and \ef{g-bd-1} also hold. This implies that 
\bes{
\bbE \left [ \nm{f-\hat{f}}^2_{L^2_{\varrho}(D)} \chi_{\nms{G-I}_2 \leq \delta} \right ] \leq \nms{g}^2_{L^2_{\varrho}(D)} +  \frac{2}{1-\delta} \bbE\left [ \nms{e}^2_{2,w} \right ] + \frac{2}{(1-\delta)^2} \sum^{n}_{l=1} \bbE \left[ |\ip{g}{\phi_l}_{\mathsf{disc},w}|^2 \right ].
}
The result now follows from Lemma \ref{lem:exp-inner-prod-perp}.
}

\prf{
[Proof of Theorem \ref{t:err-bd-exp}]
Let $E$ be the event that $\nms{G - I}_2 \leq \delta$. If $E$ occurs then
\bes{
\nm{f - \hat{f}^{\mathsf{ce}}}_{L^2_{\varrho}(D)} = \nm{f - \hat{f}}_{L^2_{\varrho}(D)},\qquad \nm{f - \hat{f}^{\mathsf{te}}}_{L^2_{\varrho}(D)} \leq \nm{f - \hat{f}}_{L^2_{\varrho}(D)}.
}
Here, the second bound follows from the facts that $f = \cT_{\sigma}(f) $ and $\cT_{\sigma}$ is a contraction in the $L^2_{\varrho}$-norm. On the other hand, if $E$ does not occur then we have
\bes{
\nm{f - \hat{f}^{\mathsf{ce}}}_{L^2_{\varrho}(D)} = \nm{f}_{L^2_{\varrho}(D)},\qquad \nm{f - \hat{f}^{\mathsf{te}}}_{L^2_{\varrho}(D)} \leq \nm{f} + \nm{ \hat{f}^{\mathsf{te}}}_{L^2_{\varrho}(D)} \leq 2 \sigma.
}
Now observe that $\bbP(E^c) \leq \epsilon$, due to the assumption on $m$ and Theorem \ref{t:alpha-beta-est}. The result now follows by the law of total expectation and Lemma \ref{l:exp-bounds-chi}.
}

\prf{
[Proof of Theorem \ref{thm:ullrich-approach}]
The weight function $w$ corresponding to \ef{opt-meas-krieg} is given by
\be{
\label{w-krieg-def}
w(x) = \left ( \frac12 + \frac14 \frac{\sum^{n}_{i=1} | \phi_i (x) |^2}{n} + \frac14 \sum^{\infty}_{l=0} \frac{v^2_l}{|I_l|} \sum_{i \in I_l} | \phi_i(x) |^2 \right )^{-1}.
}
Therefore, by \ef{Kappa-def-alt}, we have $w(x) \leq 4 / \cK(\cP)(x)$, which gives
\bes{ 
\kappa_w = \esssup_{x \sim \varrho} w(x) \cK(\cP)(x)  \leq 4 n.
}
We deduce from Theorem \ref{t:alpha-beta-est} and \ef{m-est-krieg} that $\sqrt{1-\delta} < \alpha_w \leq \beta_w \leq \sqrt{1+\delta}$ with probability at least $1-\epsilon/2$. This,   Lemma \ref{l:wLS-acc-stab} and the fact that $w(x) \leq 2$ by construction now imply that
\bes{
\nm{f - \hat{f}}_{L^2_{\varrho}(D)} \leq \inf_{p \in \cP} \left \{ \nm{f - p}_{L^{2}_{\varrho}(D)} + \frac{1}{\sqrt{1-\delta}} \nm{f - p}_{\mathsf{disc},w} \right \} + \sqrt{\frac{2}{1-\delta}} \frac{\nms{e}_{2}}{\sqrt{m}}
} 
for any $f$ and $e$, with the same probability. Next, let $c_i = \ip{f}{\phi_i}_{L^2_{\varrho}(D)}$ be the coefficients of $f$ and $p = \sum^{n}_{i=1} c_i \phi_i$ be the best approximation to $f$ from $\cP$. This gives
\be{
\label{f-krieg-bound-1} 
\nm{f - \hat{f}}_{L^2_{\varrho}(D)} \leq e_n(f) + \frac{1}{\sqrt{1-\delta}} \nm{f - p}_{\mathsf{disc},w} + \sqrt{\frac{2}{1-\delta}} \frac{\nms{e}_{2}}{\sqrt{m}}.
}
For $l = 0,1,2,\ldots$, define the matrices
\bes{
A^{(l)} = \left ( \sqrt{\frac{w(x_i)}{m}} \phi_j(x_i) \right )^{m,2^{l+1} n}_{i = 1,j=2^l n + 1} \in \bbR^{m \times 2^l n},
\qquad
c^{(l)} = \left ( c_i \right )^{2^{l+1} n}_{i=2^l n + 1} \in \bbR^{2^l n}.
}
Then
\be{
\label{f-p-expand-disc}
\nms{f - p}_{\mathsf{disc,w}} \leq \sum^{\infty}_{l=0} \nm{A^{(l)} c^{(l)}}_2 \leq \sum^{\infty}_{l=0} \nm{A^{(l)} }_2 \nm{c^{(l)}}_2 \leq \sum^{\infty}_{l=0} \nm{A^{(l)}}_2 e_{2^l n}(f),
}
where $e_{2^l n}(f)$ is as in \ef{e-def}.

We now wish to use the matrix Chernoff bound to estimate $\nm{A^{(l)}}_2$. Observe that 
\bes{
\bbP \left ( \nm{A^{(l)}}^2_2 \geq 1+t \right ) = \bbP \left ( \lambda_{\max}\left ( 
(A^{(l)})^* A^{(l)} \right ) \geq 1+t \right ).
}
As in the proof of Theorem \ref{t:alpha-beta-est}, note that $\bbE [ (A^{(l)})^* A^{(l)}  ] = I$
and
\bes{
(A^{(l)})^* A^{(l)} = \sum^{m}_{i=1} X_i,\quad \text{where }X_i = \frac1m \left ( w(x_i) \overline{\phi_j(x_i)} \phi_k(x_i) \right ) _{j,k \in I_l}.
}
The matrices $X_i \succeq 0$ and satisfy, for any $c = (c_j)_{j \in I_l}$,
\bes{
c^* X_i c = \frac{w(x_i)}{m} \left | \sum_{j \in I_l} c_j \phi_j(x_i) \right |^2 \leq \frac{w(x_i)}{m} \sum_{j \in I_l} | \phi_j(x_i) |^2 \nms{c}^2_2 .
}
Using \ef{w-krieg-def} and taking the supremum over all such $c$ with $\nms{c}_2 = 1$, we deduce that
\bes{
\lambda_{\max}(X_i) \leq \frac{4 |I_l|}{m v^2_l}.
}
Hence, the matrix Chernoff bound (Theorem \ref{t:matrix-chernoff}) gives that
\be{
\label{Al-norm-prob}
\bbP \left ( \nm{A^{(l)}}^2_2 \geq 1+t \right ) \leq |I_l| \exp \left ( - \frac{m v^2_l h(t)}{4 |I_l|} \right ),\quad \forall t \geq 0,
}
where $h(t) = (1+t) \log(1+t) - t$.
Now let $\epsilon_l = (3/\pi^2) \epsilon / (l+1)^2$, so that $\sum^{\infty}_{l=0} \epsilon_l = \epsilon / 2$. We want to choose $t = t_l$ so that $\bbP ( \nm{A^{(l)}}^2_2 \geq 1+t_l) \leq \epsilon_l $. Using \ef{Al-norm-prob} and the bound \ef{m-est-krieg}, we see that
\bes{
\bbP \left ( \nm{A^{(l)}}^2_2 \geq 1+t_l \right ) \leq | I_l| \exp \left ( - \frac{C_{\delta} n \log(4n/\epsilon) h(t_l) v^2_l}{|I_l|} \right ) \leq \epsilon_l,
}
provided
\bes{
h(t_l) \geq \frac{| I_l| \log(|I_l| / \epsilon_l) }{C_{\delta} n \log(4n/\epsilon) v^2_l }.
}
The function $h$ is increasing and $h(t) \geq h(1) t$ for $t \geq 1$. Therefore, it suffices to take
\bes{
t_l =  \frac{| I_l| \log(|I_l| / \epsilon_l) }{h(1) C_{\delta} n \log(4n/\epsilon) v^2_l }.
}
Now we recall that $|I_l| = n 2^{l}$, $\epsilon_l = (3/\pi^2) \epsilon / (l+1)^2 $ and $C_{\delta} = ((1+\delta) \log(1+\delta)-\delta))^{-1} \geq C_1 
\gtrsim 1$, to get, after some algebra,
\bes{
1+t_l \leq C \frac{2^l \log(2^{l+1} (l+1)^2)}{v^2_l}
}
for some numerical constant $C> 0$. Hence, we have shown that
\bes{
\bbP \left ( \nm{A^{(l)}}^2_2 \geq c \frac{2^l \log(2^{l+1} (l+1)^2)}{v^2_l} \right ) \leq \epsilon_l.
}
Taking the union bound and recalling that $\sum^{\infty}_{l=0} \epsilon_l = \epsilon/2$, we deduce from this and \ef{f-p-expand-disc} that
\bes{
\nms{f - p}_{\mathsf{disc,w}} \leq  C \sum^{\infty}_{l=0} 2^{l/2} \frac{\sqrt{\log(2^{l+1}(l+1)^2)}}{v_l}  e_{2^l n}(f) \leq C \sum^{\infty}_{l=0} \frac{2^{l/2} \sqrt{l+1}}{v_l} e_{2^l n}(f),
}
with probability at least $1-\epsilon/2$ and a potentially different numerical constant $C$. Now consider $e_{2^l n}(f)$. The terms $e_k(f)$ are monotonically nonincreasing in $k$. Therefore, for $l = 1,2,\ldots$ we have
\bes{
n (2^{l}-1) (e_{2^l n}(f))^p \leq e_{n+1}(f))^p + \cdots + (e_{2^l n}(f))^p \leq \sum_{k > n} (e_k(f))^p .
}
Hence
\bes{
e_{2^l n}(f) \leq C_p 2^{-l/p} \left ( \frac1n \sum_{k > n} (e_k(f))^p \right )^{1/p},\quad l = 1,2,\ldots.
}
We deduce that 
\eas{
\nms{f - p}_{\mathsf{disc,w}} & \leq C_p \left [ \frac{e_n(f)}{v_0} + \left ( \frac1n \sum_{k > n} (e_k(f))^p \right )^{1/p}  \left ( \sum^{\infty}_{l=1} \frac{2^{l(1/2-1/p)} \sqrt{l+1}}{v_l}  \right ) \right ]
 \\
 & \leq C_{p,\theta} \left [ e_n(f) +   \left ( \frac1n \sum_{k > n} (e_k(f))^p \right )^{1/p} \right ],
}
where in the final step we used the fact that $v_l = 2^{-\theta l}$ for $0 < \theta < 1/p-1/2$ to deduce that the final sum converges. Substituting this into \ef{f-krieg-bound-1} now gives the result.
}

\prf{
[Proof of Theorem \ref{thm:gen-framework}]

Much like in Lemma \ref{l:wLS-acc-stab}, if \ef{emp-nondegen} holds with $\alpha_w > 0$ then the estimator $\hat{f}$ given by \ef{general-least-squares} satisfies (see \cite[Lem.\ E.5]{adcock2023cs4ml})
\bes{
\nm{f - \hat{f}}_{\bbX} \leq \inf_{p \in \cP} \left \{ \nm{f-p}_{\bbX} + \frac{2}{\alpha_w} \nm{f-p}_{\mathsf{disc},w} \right \} + \frac{2}{\alpha_w} \nm{e}_{2,w},
} 
where, for convenience, we define $\nm{e}^2_{2,w} = \sum^{C}_{c=1} \frac{1}{m_c} \sum^{m_c}_{i=1} w_c(\theta_{ic}) \nm{e_{ic}}^2_{\bbY_c}$ and
\bes{
\nm{g}^2_{\mathsf{disc},w} = \sum^{C}_{c=1} \frac{1}{m_c} \sum^{m_c}_{c=1} w(\theta_{ic}) \nm{L_c(\theta_{ic})(g)}^2_{\bbY_c} ,\quad g \in \bbX_0.
}
Now let $E$ be the event that $\alpha_w \geq a/2$, where $a$ is as in \ef{nondegeneracy} (the choice of $1/2$ here is arbitrary), and consider the estimator $\hat{f}^{\mathsf{tr}}$. We argue similarly to the proofs of Lemma \ref{l:exp-bounds-chi} and Theorem \ref{t:err-bd-exp}. Since $\cT_{\sigma}$ is a contraction and $\sigma \geq \nm{f}_{\bbX}$, we have
\bes{
\nm{f - \hat{f}^{\mathsf{tr}}}_{\bbX} \leq \nm{f - \hat{f}}_{\bbX}\quad \text{and}\quad
\nm{f - \hat{f}^{\mathsf{tr}}}_{\bbX} \leq 2 \sigma.
}
Now fix $p \in \cP$. Then
\eas{
\bbE \left[  \nm{f - \hat{f}^{\mathsf{tr}}}^2_{\bbX} \right ] & = \bbE \left [ \nm{f - \hat{f}^{\mathsf{tr}}}^2_{\bbX} | E \right ] \bbP(E) + \bbE \left [ \nm{f - \hat{f}^{\mathsf{tr}}}^2_{\bbX} | E^c \right ]\bbP(E^c)
\\
& \leq 3 \nm{f - p}^2_{\bbX} + \frac{24}{a^2} \bbE \nm{f - p}^2_{\mathsf{disc},w} + \frac{24}{a^2} \bbE \nm{e}^2_{2,w} + 4 \sigma^2 \bbP(E^c).
}
Observe that
\eas{
\bbE \left [ \nm{f - p}^2_{\mathsf{disc},w} \right ] & = \bbE \left [ \sum^{C}_{c=1} \frac{1}{m_c} \sum^{m_c}_{i=1} w_c(\theta_{ic}) \nm{L_c(f-p)(\theta_{ic})}^2_{\bbY_c} \right ]
\\
& =  \sum^{C}_{c=1} \int_{D_c} w_c(\theta_c) \nm{L_c(f-p)(\theta_{c})}^2_{\bbY_c} \D \mu_c(\theta_c) 
\\
& = \sum^{C}_{c=1} \int_{D_c}\nm{L_c(f-p)(\theta_{c})}^2_{\bbY_c} \D \varrho_c(\theta_c)  \leq b^2 \nm{f-p}^2_{\bbX},
}
where in the last step we used nondegeneracy (Assumption \ref{ass:nondegeneracy}). We also have
\bes{
\bbE \left [ \nm{e}^2_{2,w} \right ] = \sum^{C}_{c=1} \frac{1}{m_c} \sum^{m_c}_{i=1} \nm{e_{ic}}^2_{\bbY_c} \int_{D_c} w_{c}(\theta_{c}) \D \mu_c(\theta_c) = \sum^{C}_{c=1} \frac{1}{m_c} \sum^{m_c}_{i=1} \nm{e_{ic}}^2_{\bbY_c},
}
Here we used the definition of the weight functions $w_c$ and the fact that each $\varrho_c$ is a probability measure. We deduce that
\bes{
\bbE \left [ \nm{f - \hat{f}^{\mathsf{tr}}}^2_{\bbX} \right ]\leq 3 \nm{f - p}^2_{\bbX} + \frac{24 b^2}{a^2} \nm{f - p}^2_{\bbX} + \frac{24}{a^2} \nm{e}^2_{2} + 4 \sigma^2 \bbP(E^c),
}
Hence, the result follows, provided $\bbP(E^c) \leq \epsilon$.

To show that $\bbP(E^c) \leq \epsilon$, we appeal to \cite[Thm.\ E.2]{adcock2024unified}, using conditions (b) and (c) defined therein. The remainder of the proof involves showing how to recast the setup considered in \S \ref{ss:general-setup} as a special case of that considered in \cite{adcock2024unified}. Let $m = m_1+\cdots + m_C$. Now, for $c = 1,\ldots,C$, let $\cA^{(c)}$ be the distribution of operators in $\cB(\bbX_0,\bbY_c)$ defined by $A^{(c)} \sim \cA^{(c)}$ if
\bes{
A^{(c)}(f) = \sqrt{m/m_c} \sqrt{w_c(\theta_c)} L_c(\theta_c)(f),\quad \text{where }\theta_c \sim \mu_c.
}
Following \cite[Ex.\ 2.5]{adcock2024unified}, we define $\{\cA_i \}^{m}_{i=1}$ by $\cA_1 = \cdots = \cA_{m_1} = \cA^{(1)}$ and $\cA_i = \cA^{(c)}$ if $m_1+\cdots + m_{c-1} < i \leq m_1 + \cdots + m_c$ for $c = 2,\ldots,C$. Doing this, the setup of \S \ref{ss:general-setup} becomes a special case of \cite{adcock2024unified}. In particular, nondegeneracy in the sense of \cite[eqn.\ (1.1)]{adcock2024unified} is implied by \ef{nondegeneracy}. 

We now apply \cite[Thm.\ E.2]{adcock2024unified}, and specifically, parts (b) and (c), with $\bbU = \cP'$ and $\delta = 1/2$. This implies that \ef{emp-nondegen} holds, provided
\be{
\label{m-cond-gen-1}
m \gtrsim a^{-2} \cdot \Phi(S(\cP'') ; \bar{\cA}) \cdot (\log(2 d / \epsilon) + n ),
}
or, with $\cQ$ as in Assumption \ref{ass:UofS}(b), 
\be{
\label{m-cond-gen-2}
m \gtrsim a^{-2} \cdot \Phi(S(\cQ) ; \bar{\cA}) \cdot \log(2 d n/ \epsilon) . 
}
where $\Phi$ is the so-called \textit{variation}, as defined in \cite[\S 3.1]{adcock2024unified} and $S(\cU) = \{ u / \nm{u}_{\bbX} : u \in \cU,\ u \neq 0 \}$ for any $\cU \subseteq \bbX_0$, $\cU \neq \{0\}$. Consider any such set $\cU$. Using this and the definition of $\bar{\cA}$, we see that
\bes{
\Phi(S(\cU) ; \bar{\cA}) = \max_{c=1,\ldots,C} \Phi(S(\cU) ; \cA^{(c)} ) =  \max_{c=1,\ldots,C} \left \{ \frac{m}{m_c} \esssup_{\theta_c \sim \varrho_c} \sup_{u \in \cU \backslash \{ 0 \}} w_c(\theta_c) \nm{L_c(\theta_c)(u / \nm{u}_{\bbX})}^2_{\bbY_c} \right \}.
}
Since $L_c(\theta_c)$ is linear, we deduce that
\bes{
\Phi(S(\cU) ; \bar{\cA}) = \max_{c=1,\ldots,C} \left \{ \frac{m}{m_c} \kappa_{w_c}(\cU ; L_c) \right \},
}
where $\kappa_{w_c}$ is as in \ef{kappa-w-def-general}. Using this and setting, respectively, $\cU = \cP''$ or $\cU = \cQ$ we see that \ef{m-cond-gen-1} is equivalent to \ef{mc-cond-1} and \ef{m-cond-gen-2} is equivalent to \ef{mc-cond-2}. The result now follows.
}

\end{document}